\documentclass[acmtog]{acmart}
\usepackage{multirow}
\usepackage{booktabs}
\usepackage{graphicx}
\usepackage{multirow}
\AtBeginDocument{%
  }

\setcopyright{acmlicensed}
\copyrightyear{2026}
\acmYear{2026}
\setcopyright{cc}
\setcctype{by}
\acmConference[SIGGRAPH Conference Papers '26]{Special Interest Group on Computer Graphics and Interactive Techniques Conference Conference Papers}{July 19--23, 2026}{Los Angeles, CA, USA}
\acmBooktitle{Special Interest Group on Computer Graphics and Interactive Techniques Conference Conference Papers (SIGGRAPH Conference Papers '26), July 19--23, 2026, Los Angeles, CA, USA}
\acmDOI{10.1145/3799902.3811150}
\acmISBN{979-8-4007-2554-8/2026/07}

\acmSubmissionID{873}

\usepackage[dvipsnames]{xcolor}
\usepackage{enumitem}

\usepackage{fancyvrb}

\definecolor{myblue}{RGB}{68, 114, 196}
\definecolor{myorange}{RGB}{237, 125, 49}
\definecolor{mygreen}{RGB}{112, 173, 71}

\usepackage{tcolorbox}
\tcbuselibrary{breakable,skins}
\usepackage{xcolor}

\tcbset{
  promptbox/.style={
    breakable,
    enhanced,
    colback=gray!6,
    colframe=black!55,
    fonttitle=\bfseries\scriptsize,
    title filled=false,
    left=4pt, right=4pt, top=4pt, bottom=4pt,
  }
}

\begin{document}

%%
%% The "title" command has an optional parameter,
%% allowing the author to define a "short title" to be used in page headers.
\title{Stroke of Surprise: Progressive Semantic Illusions in Vector Sketching}

%%
%% The "author" command and its associated commands are used to define
%% the authors and their affiliations.
%% Of note is the shared affiliation of the first two authors, and the
%% "authornote" and "authornotemark" commands
%% used to denote shared contribution to the research.
\author{Huai-Hsun Cheng}
\affiliation{%
  \institution{National Yang Ming Chiao Tung University}
  \country{Taiwan}}
\email{huaish.cs13@nycu.edu.tw}

\author{Siang-Ling Zhang}
\affiliation{%
  \institution{National Yang Ming Chiao Tung University}
  \country{Taiwan}}
\email{siang1105.cs13@nycu.edu.tw}

\author{Yu-Lun Liu}
\affiliation{%
  \institution{National Yang Ming Chiao Tung University}
  \country{Taiwan}}
\email{yulunliu@cs.nycu.edu.tw}

%%
%% By default, the full list of authors will be used in the page
%% headers. Often, this list is too long, and will overlap
%% other information printed in the page headers. This command allows
%% the author to define a more concise list
%% of authors' names for this purpose.
\renewcommand{\shortauthors}{Cheng et al.}

%%
%% The abstract is a short summary of the work to be presented in the
%% article.
\begin{abstract}
  Visual illusions traditionally rely on spatial manipulations such as multi-view consistency. In this work, we introduce Progressive Semantic Illusions, a novel vector sketching task where a single sketch undergoes a dramatic semantic transformation through the sequential addition of strokes. We present Stroke of Surprise, a generative framework that optimizes vector strokes to satisfy distinct semantic interpretations at different drawing stages. The core challenge lies in the ``dual-constraint'': initial prefix strokes must form a coherent object (e.g., a duck) while simultaneously serving as the structural foundation for a second concept (e.g., a sheep) upon adding delta strokes. To address this, we propose a sequence-aware joint optimization framework driven by a dual-branch Score Distillation Sampling (SDS) mechanism. Unlike sequential approaches that freeze the initial state, our method dynamically adjusts prefix strokes to discover a ``common structural subspace'' valid for both targets. Furthermore, we introduce a novel Overlay Loss that enforces spatial complementarity, ensuring structural integration rather than occlusion. Extensive experiments demonstrate that our method significantly outperforms state-of-the-art baselines in recognizability and illusion strength, successfully expanding visual anagrams from the spatial to the temporal dimension.
  Project page: \url{https://stroke-of-surprise.github.io/}
\end{abstract}

%%
%% The code below is generated by the tool at http://dl.acm.org/ccs.cfm.
%% Please copy and paste the code instead of the example below.
%%


% Vector sketching
% Visual illusions
% Generative AI
% Score Distillation Sampling (SDS)
% Diffusion models
% Progressive generation

%%
%% Keywords. The author(s) should pick words that accurately describe
%% the work being presented. Separate the keywords with commas.
% \keywords{Vector sketching, Visual illusions, Diffusion models, Progressive generation}
%% A "teaser" image appears between the author and affiliation
%% information and the body of the document, and typically spans the
%% page.
\begin{teaserfigure}
  \includegraphics[width=\textwidth]{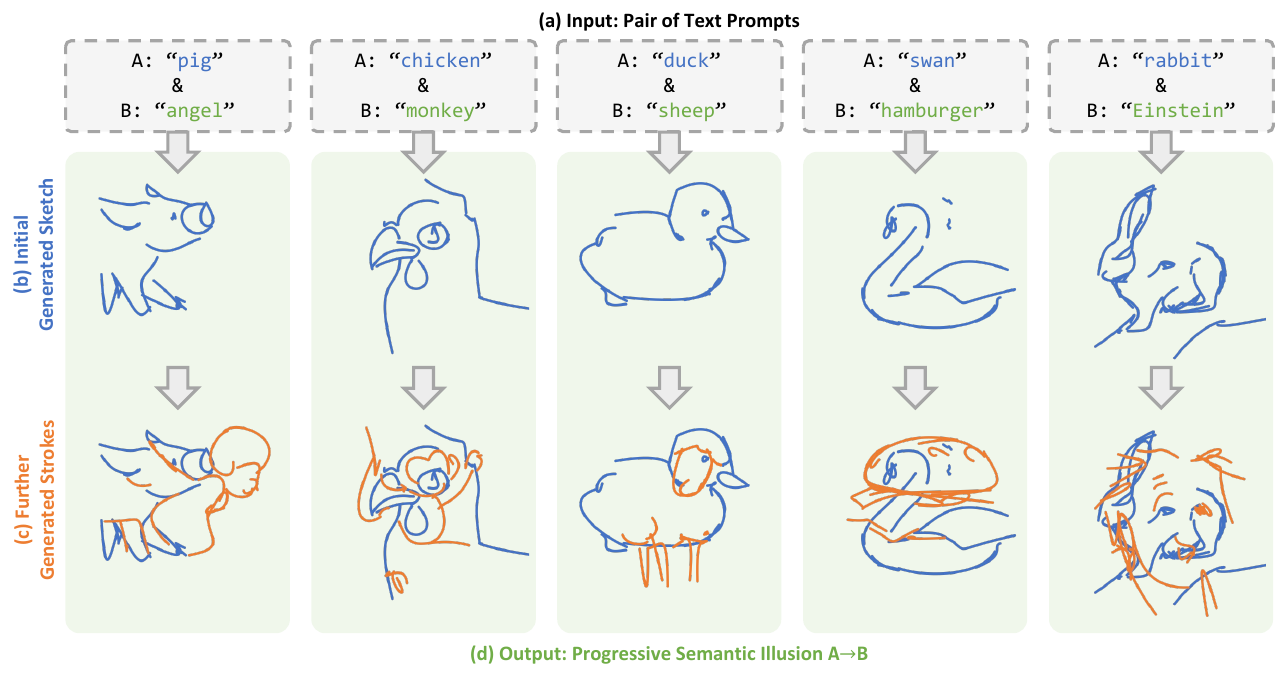}
  \caption{
  \textbf{Progressive semantic illusions from text.} Given a pair of text prompts (a), our method generates a vector sketch that evolves over time. The \textcolor{myblue}{initial generated sketch} (b) depicts the first concept (e.g., \texttt{``pig''}). By adding \textcolor{myorange}{further generated strokes} (c), the drawing is transformed into a totally different object (e.g., \texttt{``angel''}). This creates a \textit{\textcolor{mygreen}{Stroke of Surprise}}: the process \textbf{subverts the viewer's expectation} of the initial concept, triggering a dramatic \textbf{semantic reversal} as the final strokes re-contextualize the entire composition.
  % Our method generates an \textcolor{myblue}{initial sketch} (a) and adds \textcolor{myorange}{further generated strokes} to create an illusion (b) that is perceived as a totally different concept (e.g., a pig becomes an angel).
  }
  \label{fig:teaser}
\end{teaserfigure}

%%
%% This command processes the author and affiliation and title
%% information and builds the first part of the formatted document.
\maketitle

\section{Introduction}

% Visual illusions traditionally exploit spatial ambiguities, requiring viewers to change viewpoints to perceive hidden meanings (e.g., ``Visual Anagrams''~\cite{geng2024visualanagrams}). In this work, we introduce a new dimension to this artistic interplay: \emph{time}. We propose \textbf{Progressive Semantic Illusions}, a novel vector sketching task where the drawing process itself drives semantic transformation. \huaish{Sparse line drawings are uniquely suited for temporal semantic transformation: their incompleteness invites Gestalt closure~\cite{wagemans2012gestalt}, enabling the visual system to re-interpret the same strokes as new ones arrive.} As illustrated in Fig.~\ref{fig:teaser}, our method generates a coherent initial sketch (e.g., ``a pig'') that is subsequently re-contextualized by additional strokes into a distinct concept (e.g., ``an angel''). This ``Stroke of Surprise'' subverts expectations, achieving a perceptual shift through sequential stroke accumulation rather than spatial manipulation.

Visual illusions traditionally exploit spatial ambiguities, requiring viewpoint changes to reveal hidden meanings~\cite{geng2024visualanagrams}. We introduce a new dimension: \emph{time}. We propose \textbf{Progressive Semantic Illusions}, where the drawing process itself drives semantic transformation. Sparse line drawings are uniquely suited: their incompleteness invites Gestalt closure~\cite{wagemans2012gestalt}, letting the visual system re-interpret existing strokes as new ones arrive. As shown in Fig.~\ref{fig:teaser}, our method generates an initial sketch (e.g., ``a pig'') that additional strokes re-contextualize into a distinct concept (e.g., ``an angel''), achieving perceptual shift through sequential accumulation rather than spatial manipulation.

Sketch generation has evolved from category-specific RNNs~\cite{ha2017neural} to open-vocabulary models leveraging CLIP~\cite{radford2021learning} and diffusion priors~\cite{rombach2022ldm}. Methods like CLIPasso~\cite{vinker2022clipasso} and VectorFusion~\cite{jain2023vectorfusion} utilize differentiable rasterization for high-fidelity sketching, while sequential approaches like SketchAgent~\cite{vinker2025sketchagent} and SketchDreamer~\cite{qu2023sketchdreamer} mimic step-by-step human drawing. Regarding illusions, Visual Anagrams~\cite{geng2024visualanagrams} and ShadowDraw~\cite{luo2025shadowdraw} explore multi-view effects via diffusion. However, these prior works focus on static pixel representations or spatial rearrangements, leaving the challenge of \emph{temporal} semantic transformation in vector graphics unexplored.

\begin{figure}
  \includegraphics[width=\columnwidth]{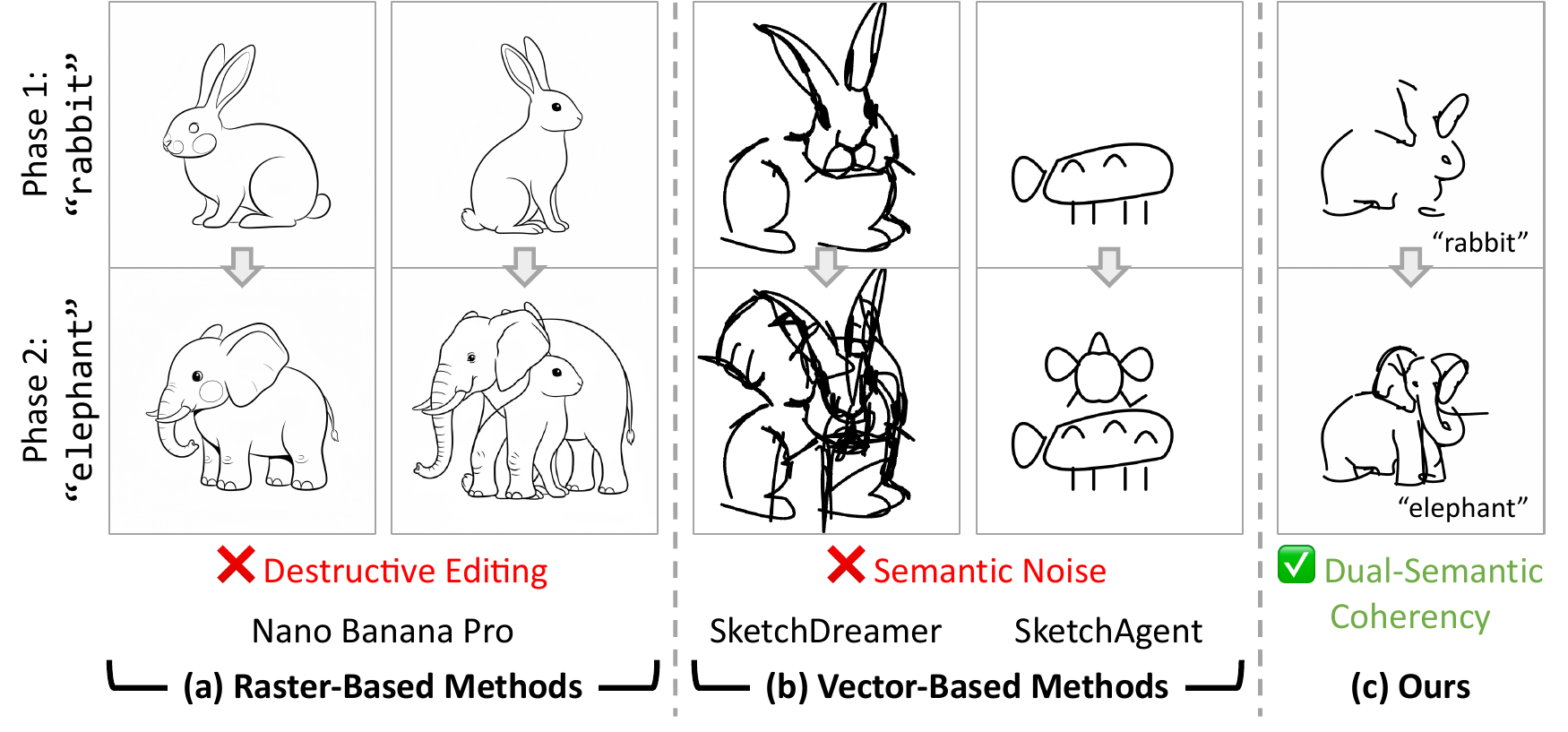}
  \caption{
  \textbf{Challenges in progressive illusion sketching.}
(a) Raster-based methods (e.g., Nano Banana Pro) rely on \textcolor{red}{\textbf{destructive editing}}, modifying the initial structure to fit the final target and thus violating the progressive constraint.
(b) Vector-based baselines (e.g., SketchDreamer~\cite{qu2023sketchdreamer} or SketchAgent~\cite{vinker2025sketchagent}) employ a greedy strategy, where specific Phase 1 details become \textcolor{red}{\textbf{semantic noise}} or clutter in Phase 2.
(c) Ours achieves \textcolor{mygreen}{\textbf{dual-semantic coherency}} by jointly optimizing for a common structural subspace, ensuring the initial strokes are valid building blocks for both interpretations (e.g., \texttt{``rabbit''} $\rightarrow$ \texttt{``elephant''}).
  % \textbf{Limitations of existing methods for dual-constraint illusion sketch.} Each column shows sketches generated by different methods, with Phase 1 (top) and Phase 2 (bottom) for each approach. (a) Raster-based methods (e.g., Gemini) lack explicit stroke ordering, leading to loss of Phase 1 structure during Phase 2 generation. (b, c) Vector-based methods (e.g., SketchDreamer~\cite{qu2023sketchdreamer}, SketchAgent~\cite{vinker2025sketchagent}) explicitly maintain stroke sequences but optimize for single semantic targets, resulting in structural inconsistency when extending sketches across phases. (d) Our method employs a sequence-aware optimization framework with shared stroke parameters across multiple semantic phases, achieving coherent cross-phase transformations.
  }
  \label{fig:motivation}
\end{figure}

Generating progressive illusions presents a unique ``Dual-Constraint'': early strokes must depict object ``A'' while simultaneously functioning as the structural foundation for object ``B'' (Fig.~\ref{fig:motivation}). Existing methods fail to address this additive nature. Raster-based models (e.g., Nano Banana Pro) rely on \emph{destructive editing}, overwriting initial pixels and violating the progressive constraint. Conversely, sequential vector models (e.g., SketchAgent) employ a \emph{greedy strategy}, optimizing strokes solely for ``A''. This renders the fixed prefix as \emph{semantic noise} when extending to ``B'', resulting in clutter. Crucially, these baselines fail to find a \textbf{''Common Subspace''}, which is a shared geometric configuration valid for both semantic interpretations.

\begin{figure*}[t]
  \centering
  \includegraphics[width=\textwidth]{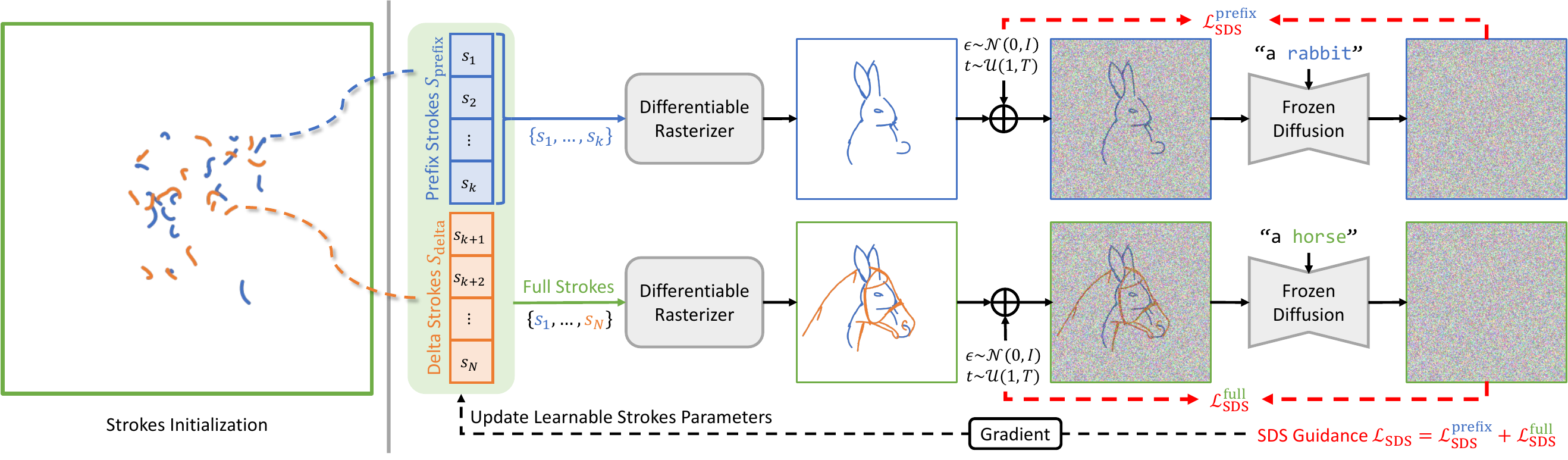}
  \caption{\textbf{Pipeline overview.}
  Our method optimizes a set of learnable stroke parameters, which are divided into \textcolor{myblue}{prefix strokes $S_\text{prefix}$} and \textcolor{myorange}{delta strokes $S_\text{delta}$}. The optimization process involves two parallel branches. In the top branch, only the prefix strokes are rendered by a differentiable rasterizer to create a partial sketch (e.g., a rabbit). This sketch is then guided by a pre-trained, frozen text-to-image diffusion model using a prompt corresponding to the prefix (``\texttt{a rabbit}''), resulting in the prefix SDS loss $\mathcal{L}_{\text{SDS}}^{\text{prefix}}$. In the bottom branch, the \textcolor{mygreen}{full set of strokes} is rendered to create the complete sketch (e.g., a horse). This is guided by the same diffusion model using a prompt for the full object (``\texttt{a horse}''), resulting in the full SDS loss  $ \mathcal{L}_{\text{SDS}}^{\text{full}}$. The total SDS guidance loss is the sum of these two terms $\mathcal{L}_{\text{SDS}} = \mathcal{L}_{\text{SDS}}^{\text{prefix}} + \mathcal{L}_{\text{SDS}}^{\text{full}}$. Gradients from this total loss are backpropagated to update all learnable stroke parameters.
  }
\label{fig:pipeline}
\end{figure*}

To overcome these limitations, we propose \textbf{Stroke of Surprise}, a sequence-aware joint optimization framework designed to discover this common structural subspace (Fig.~\ref{fig:pipeline}). Unlike sequential approaches, we optimize parameters for both the prefix (Object ``A'') and full phase (Object ``B'') simultaneously using a dual-branch Score Distillation Sampling (SDS) mechanism. This guidance ensures prefix strokes are recognizable as the initial concept yet ``primed'' for re-interpretation. Furthermore, we introduce a geometric \emph{Overlay Loss} to enforce spatial complementarity and prevent occlusion. This enables delta strokes to structurally integrate with and re-contextualize the prefix. For example, it can transform pig ears into angel wings, creating a seamless illusion.

Our main contributions are summarized as follows:
\begin{itemize}[leftmargin=*]
    \item \textbf{Task:} We introduce \textbf{Progressive Semantic Illusion}, extending visual illusions from the spatial to the \emph{temporal} dimension. This task requires a single vector sketch to reveal distinct semantic interpretations through progressive stroke accumulation.
   
    % \item \textbf{Method:} We propose a \textbf{sequence-aware joint optimization framework} that optimizes shared stroke parameters under simultaneous semantic constraints and a novel Overlay Loss, ensuring the prefix strokes form a robust structural foundation for the final illusion.
    \item \textbf{Method:} 
    % We formalize this task as a constrained optimization over shared Bézier stroke parameters, enabling joint optimization that discovers a \textbf{Common Structural Subspace}. We propose a novel \textbf{Overlay Loss}, a geometric regularizer that enforces spatial complementarity between prefix and delta strokes, preventing perceptual crowding and ensuring structural integration over occlusion. We further introduce a \textbf{VLM-based filtering and ranking pipeline} for systematic candidate selection.
    We formalize this as constrained optimization over shared Bézier parameters, enabling joint discovery of a \textbf{Common Structural Subspace}. A novel \textbf{Overlay Loss} enforces spatial complementarity between prefix and delta strokes, preventing crowding and ensuring integration over occlusion. We further introduce a \textbf{VLM-based filtering and ranking pipeline} for candidate selection.
    
    % \item \textbf{Insight \& Scalability}: We demonstrate that joint optimization identifies a ``\textbf{Common Subspace}'' that resolves conflicts between early semantic clarity and later structural integration. Our method scales to multi-phase illusions (``A'' $\rightarrow$ ``B'' $\rightarrow$ ``C'') and significantly outperforms baselines in recognizability and coherence.
    \item \textbf{Evaluation \& Scalability:} 
    % Extensive experiments and user studies validate that our method significantly outperforms state-of-the-art baselines in recognizability and illusion strength. Our framework naturally generalizes to $K$-phase illusions (``A'' $\rightarrow$ ``B'' $\rightarrow$ ``C'') and alternative representations including B-spline curves, colored strokes, and general vector graphics.
    Experiments and user studies show our method significantly outperforms baselines in recognizability and illusion strength. Our framework generalizes to $K$-phase illusions (``A'' $\rightarrow$ ``B'' $\rightarrow$ ``C'') and alternative representations including B-splines, colored strokes, and general vector graphics.
\end{itemize}

\section{Related Work}
\label{sec:related_work}

\paragraph{Generative Vector Graphic Synthesis.}
Early sketch synthesis relied on category-specific corpora~\cite{eitz2012humans,jongejan2016quickdraw,sangkloy2016sketchy} with a fixed vocabulary. CLIP~\cite{radford2021learning} lifted this constraint. CLIPDraw~\cite{frans2022clipdraw} and CLIPasso~\cite{vinker2022clipasso} optimize Bézier curves~\cite{bezier1968,decasteljau1959} through a differentiable rasterizer~\cite{li2020differentiable} against CLIP similarity. This approach later extended to scenes~\cite{vinker2023clipascene}, though global image-text alignment lacks dense structural supervision.
Score Distillation Sampling (SDS)~\cite{poole2022dreamfusion} replaces this with per-pixel diffusion gradients. VectorFusion~\cite{jain2023vectorfusion} first ported SDS to SVGs. DiffSketcher~\cite{xing2023diffsketcher} initializes strokes from cross-attention. SVGDreamer~\cite{xing2024svgdreamer} decomposes prompts into semantic components, and SketchDreamer~\cite{qu2023sketchdreamer} adds interactive prompting. The objective itself has been refined by ProlificDreamer's variational particles~\cite{wang2023prolificdreamer}, LucidDreamer's interval matching~\cite{liang2024luciddreamer}, and SDI's reparametrized DDIM~\cite{lukoianov2024ddim}.
Others bypass optimization entirely. SwiftSketch~\cite{arar2025swiftsketch} predicts strokes feed-forward. DeepSVG~\cite{carlier2020deepsvg} learns a hierarchical SVG latent. IconShop~\cite{wu2023iconshop} and StarVector~\cite{rodriguez2025starvector} autoregressively decode SVG tokens. A parallel line replaces discrete control points with implicit fields~\cite{reddy2021im2vec,thamizharasan2024nivel}. NeuralSVG~\cite{polaczek2025neuralsvg} combines these with SDS.
All of these treat drawing as a \emph{single} static target. Our setting requires the same strokes to satisfy two interpretations at different completion stages, a constraint no static-target formulation addresses.
\paragraph{Sequential Sketch Generation.}
Sequential methods model drawing as a temporal process. DRAW~\cite{gregor2015draw} introduced iterative glimpses, and SketchRNN~\cite{ha2017neural} adapted this to stroke sequences with an LSTM, later replaced by Transformers for longer-range modeling~\cite{ribeiro2020sketchformer}.
For stroke geometry, BézierSketch~\cite{das2020beziersketch} autoregressively emits parametric curves rather than polylines. DoodleFormer~\cite{bhunia2022doodleformer} decouples coarse layout from fine detail. SketchKnitter~\cite{wang2023sketchknitter} replaces autoregressive decoding with parallel diffusion. Complementary work tackles partial-input completion~\cite{liu2019sketchgan,su2020sketchhealer} and stroke-level hierarchical editing~\cite{zang2025hierarchical}. More recently, SketchAgent~\cite{vinker2025sketchagent} drives stroke generation through LLM dialogue. While Chat2SVG~\cite{wu2025chat2svg} and LLM4SVG~\cite{xing2025llmsvg} prompt LLMs to emit SVG code directly.
Despite their temporal formulation, all commit each stroke greedily to a \emph{single} target. Once emitted, a stroke is frozen. As our ablation shows (Fig.~\ref{fig:ablation-optimization}), this traps the prefix in a local minimum incompatible with a second concept. Our joint optimization instead lets prefix strokes shift under dual semantic pressure.
\paragraph{Sketch Perception and Visual Illusions.}
Line drawings are cognitively robust. Gestalt closure and figure-ground segregation~\cite{wagemans2012gestalt} let viewers complete fragmentary contours~\cite{biederman1987geon,kanizsa1979}. Sparse sketches reliably trigger recognition in both humans~\cite{cavanagh2005artist,fan2023drawing,eitz2012humans} and networks~\cite{yu2017sketchanet}. This makes them a natural substrate for progressive illusions.
Computational illusions have almost exclusively exploited \emph{spatial} manipulation. Shadow art~\cite{mitra2009shadowart} casts different silhouettes from distinct lighting. Hybrid images~\cite{oliva2006hybrid} interleave frequency bands that change perception with viewing distance. Wire art~\cite{hsiao2018wireart} renders different 2D projections from different angles. Anamorphic sculptures~\cite{pratt2023anamorphic,wu2022mirror} or warped canvases~\cite{debnath2025rasp,chang2025lookingglass} reveal hidden images only under specific viewpoints.
Diffusion priors extend this paradigm. Visual Anagrams~\cite{geng2024visualanagrams} averages scores across flips and rotations. Factorized Diffusion~\cite{geng2024factorized} splits the score into frequency bands depicting different concepts. PTDiffusion~\cite{gao2025ptdiffusion} transfers spectral phase between prompts. Illusion3D~\cite{feng2024illusion3d} enforces 3D-viewpoint consistency. Diffusion Illusions~\cite{burgert2024diffusionillusions} adds fabrication constraints. Images that Sound~\cite{chen2024imagessound} jointly satisfies visual and spectrogram interpretations. AmbiGen~\cite{zhao2023ambigen} produces rotation-readable ambigrams.
Every one of these relies on a \emph{symmetric} spatial transform, like flip, rotate, reproject, that swaps one complete image for another. Ours is \emph{asymmetric and additive}: the prefix is a strict geometric subset of the final drawing, not a transformed counterpart. Our dual-branch SDS and Overlay Loss target precisely this temporal constraint.

% \yulunliu{Are we going to have one or two algorithm blocks here in the method section?}

\section{Method}

Progressive illusions require prefix strokes to depict an initial object while forming the structural basis for a final one. We propose a joint optimization framework via multi-branch Score Distillation Sampling to discover a common structural subspace valid for both interpretations. Prefix strokes receive simultaneous gradients to satisfy dual roles, while an overlay loss enforces spatial separation, ensuring structural integration rather than occlusion.

% Generating progressive illusion sketches requires early strokes to simultaneously depict one coherent object while forming essential structural components of an entirely different final object. Our key insight is that joint optimization under multiple semantic objectives discovers stroke configurations occupying a common structural subspace supporting both interpretations. We optimize all stroke subsets jointly through a multi-branch Score Distillation Sampling framework where prefix strokes receive gradients from multiple branches simultaneously to satisfy dual interpretative roles. An overlay loss enforces spatial separation between stroke subsets, promoting structural integration rather than occlusion.

\subsection{Progressive Semantic illusion in Vector Form}
We partition a set of learnable Bézier strokes $S$ into disjoint subsets: prefix $S_{\text{prefix}} = \{s_1, \ldots, s_k\}$ and delta $S_{\text{delta}} = \{s_{k+1}, \ldots, s_N\}$. The progressive illusion requires $S_{\text{prefix}}$ to depict the initial concept $p_1$, while the full sketch $S_{\text{full}} = S$ depicts the target $p_2$, achieved by delta strokes recontextualizing the prefix. We optimize stroke parameters $\theta$ such that the rasterized outputs $\mathcal{R}(S_{\text{prefix}}; \theta)$ and $\mathcal{R}(S_{\text{full}}; \theta)$ align with $p_1$ and $p_2$, respectively. The core challenge lies in discovering configurations where prefix strokes meaningfully serve both semantic interpretations.

\subsection{Joint Optimization Pipeline}
We employ a dual-branch strategy to simultaneously refine both stroke subsets (Fig.~\ref{fig:pipeline}). Unlike sequential methods, our pipeline coordinates semantic objectives via parallel guidance on shared learnable parameters $\theta$. We initialize $N$ strokes near the canvas center, partitioning them into $S_{\text{prefix}}$ (first $k$) and $S_{\text{delta}}$ (remaining).
At each iteration, the prefix branch renders $I_{\text{prefix}} = \mathcal{R}(S_{\text{prefix}}; \theta)$. We apply the gradient of the Score Distillation Sampling (SDS) loss conditioned on $p_1$:
\begin{equation} \small
\nabla_\theta \mathcal{L}_{\text{SDS}}^{\text{prefix}} = \left[ w(t) \left( \epsilon_\phi(z_t, t, p_1) - \epsilon \right) \frac{\partial z_t}{\partial \theta} \right],
\end{equation}
where $z_t$ is the noised latent, $\epsilon_\phi$ the noise predictor, and $w(t)$ a weighting function.

Simultaneously, the full branch renders $I_{\text{full}} = \mathcal{R}(S_{\text{full}}; \theta)$ conditioned on $p_2$, yielding $\nabla_\theta \mathcal{L}_{\text{SDS}}^{\text{full}}$. We combine these gradients as
\begin{equation}
\nabla_\theta \mathcal{L}_{\text{SDS}} = \nabla_\theta \mathcal{L}_{\text{SDS}}^{\text{prefix}} + \nabla_\theta \mathcal{L}_{\text{SDS}}^{\text{full}}.
\end{equation}
This ensures prefix strokes receive simultaneous gradients from both targets, satisfying dual roles, while delta strokes optimize to complement them. To prevent delta strokes from merely occluding the prefix, which is a common issue with pure semantic guidance, we introduce an \emph{overlay loss} that penalizes spatial overlap to enforce structural integration.

\begin{figure}[t]
  \includegraphics[width=\columnwidth]{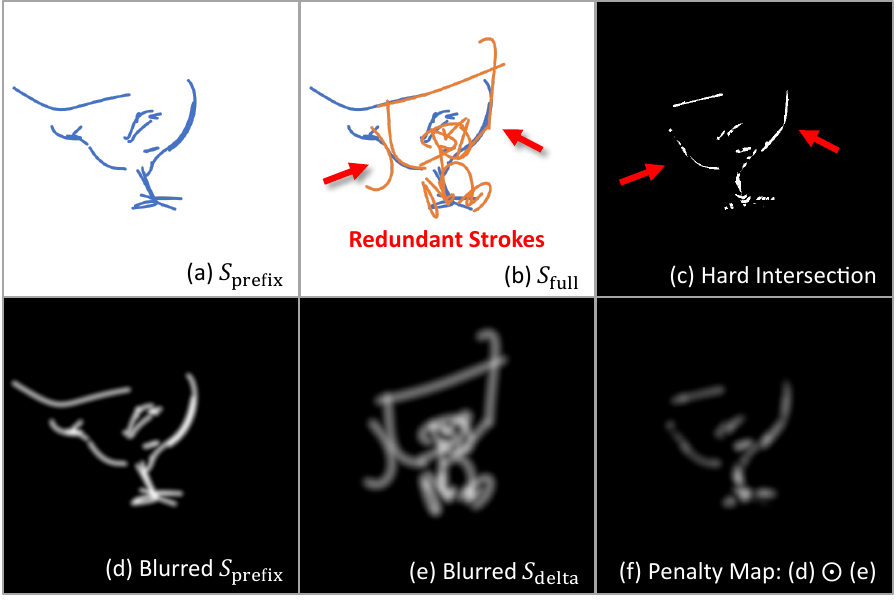}
  \caption{
    \textbf{Motivation and formulation of the overlay loss.}
% \textbf{(\emph{Top}) Motivation}: Without constraints, redundant strokes (b) occlude the prefix. Hard intersection (c) allows strokes to be placed arbitrarily close, causing crowding.
% \textbf{(\emph{Bottom}) Formulation}: We compute a \textbf{soft overlay loss} (f) from blurred maps (d, e). The blur expands the penalty region to create a \textbf{spatial buffer}, forcing new strokes to \textbf{maintain sufficient distance} from the prefix to ensure visual clarity and separation.
    \textbf{(\emph{Top}) Motivation}: (a)~The prefix sketch $S_\text{prefix}$ (\textcolor{myblue}{blue}, e.g., ``chicken'') is the structural foundation. Without spatial constraints, delta strokes (\textcolor{myorange}{orange}) cause redundant occlusions (arrows in~(b)); the hard intersection map~(c) highlights severely crowded regions. \textbf{(\emph{Bottom}) Formulation}: The \textbf{soft overlay loss}~(f) is the normalized inner product of Gaussian-blurred maps of $S_\text{prefix}$~(d) and $S_\text{delta}$~(e). Blurring creates a \textbf{spatial buffer} beyond stroke boundaries, enforcing minimum separation and structural complementarity.
    % \textbf{(\emph{Top}) Motivation}: (a)~The prefix sketch $S_\text{prefix}$ (shown in blue, e.g., ``chicken'') serves as the structural foundation. Without spatial constraints, delta strokes (shown in orange) accumulate atop the prefix, producing redundant occlusions as indicated by the arrows in~(b). The hard intersection map~(c) reveals the corresponding regions of severe spatial crowding, where delta strokes are placed with no minimum separation from the prefix.
    % \textbf{(\emph{Bottom}) Formulation}: We compute a \textbf{soft overlay loss}~(f) as the normalized inner product of Gaussian-blurred maps of $S_\text{prefix}$~(d) and $S_\text{delta}$~(e). The blurring expands the penalty region beyond stroke boundaries, creating a \textbf{spatial buffer} that enforces a minimum separation between prefix and delta strokes, ensuring structural complementarity rather than destructive occlusion.
  }
  \label{fig:overlay}
\end{figure}

\subsection{Overlay Loss for Spatial Coordination}
% While semantic guidance aligns sketches with their prompts, it provides no mechanism to prevent spatial redundancy. As shown in Fig. \ref{fig:overlay}(b), without spatial constraints, delta strokes frequently overlay existing prefix strokes, creating redundant marks that degrade visual clarity. We introduce an overlay loss that penalizes intersection between stroke subsets, ensuring delta strokes occupy complementary spatial regions.
% We render prefix and delta strokes separately to obtain $I_{\text{prefix}} = \mathcal{R}(S_{\text{prefix}}; \theta)$ and $I_{\text{delta}} = \mathcal{R}(S_{\text{delta}}; \theta)$. To capture spatial proximity beyond hard intersection (Fig. \ref{fig:overlay}(c)), we apply Gaussian blur with kernel $G_{\sigma}$, producing blurred maps (Fig. \ref{fig:overlay}(d,e)):
% \begin{equation}
% \tilde{I}_{\text{prefix}} = G_{\sigma} * I_{\text{prefix}}, \quad \tilde{I}_{\text{delta}} = G_{\sigma} * I_{\text{delta}}.
% \end{equation}
% The blur creates a spatial buffer around each stroke, discouraging excessive proximity. We compute the normalized overlap (Fig. \ref{fig:overlay}(f)):
% \begin{equation}
% \mathcal{L}_{\text{overlay}} = \frac{2 \langle \tilde{I}_{\text{prefix}}, \tilde{I}_{\text{delta}} \rangle}{\|\tilde{I}_{\text{prefix}}\|_1 + \|\tilde{I}_{\text{delta}}\|_1},
% \end{equation}
% where $\langle \cdot, \cdot \rangle$ denotes inner product and $\|\cdot\|_1$ denotes L1 norm.
Semantic guidance alone fails to prevent spatial redundancy, often causing delta strokes to clutter prefix strokes (Fig. \ref{fig:overlay}(b)). We introduce an \emph{overlay loss} to enforce spatial complementarity. We render stroke subsets separately and apply Gaussian blur $G_{\sigma}$ to create soft spatial buffers ($\tilde{I}_{\text{prefix}}, \tilde{I}_{\text{delta}}$), as shown in Fig. \ref{fig:overlay}(d,e). We then compute the normalized overlap:
\begin{equation} \small
\mathcal{L}_{\text{overlay}} = \frac{2 \langle \tilde{I}_{\text{prefix}}, \tilde{I}_{\text{delta}} \rangle}{\|\tilde{I}_{\text{prefix}}\|_1 + \|\tilde{I}_{\text{delta}}\|_1},
\end{equation}
where $\langle \cdot, \cdot \rangle$ denotes the inner product over pixel space.

% While semantic guidance aligns sketches with their prompts, it provides no mechanism to prevent spatial redundancy. As shown in Fig. \ref{fig:overlay}(b), without spatial constraints, delta strokes frequently overlay existing prefix strokes, creating redundant marks that degrade visual clarity. We introduce an overlay loss that penalizes intersection between stroke subsets, ensuring delta strokes occupy complementary spatial regions.
% We render prefix and delta strokes separately to obtain $I_{\text{prefix}}$ and $I_{\text{delta}}$. To capture spatial proximity beyond hard intersection (Fig. \ref{fig:overlay}(c)), we apply Gaussian blur $G_{\sigma}$ to both rendered images, producing blurred maps $\tilde{I}_{\text{prefix}}$ and $\tilde{I}_{\text{delta}}$ (Fig. \ref{fig:overlay}(d,e)). The blur creates a spatial buffer around each stroke, discouraging excessive proximity. We then compute the normalized overlap as:
% \begin{equation} \small
% \mathcal{L}_{\text{overlay}} = \frac{2 \langle \tilde{I}_{\text{prefix}}, \tilde{I}_{\text{delta}} \rangle}{\|\tilde{I}_{\text{prefix}}\|_1 + \|\tilde{I}_{\text{delta}}\|_1},
% \end{equation}
% where $\langle \cdot, \cdot \rangle$ denotes inner product and $\|\cdot\|_1$ denotes L1 norm (Fig. \ref{fig:overlay}(f)).

This constraint promotes structural integration and smoother semantic transitions, ensuring prefix strokes serve as essential components rather than being obscured. The final objective is:
\begin{equation} \small
\mathcal{L} = \mathcal{L}_{\text{SDS}} + \lambda_{\text{overlay}} \mathcal{L}_{\text{overlay}},
\end{equation}
where $\lambda_{\text{overlay}}$ weights the penalty. Gradients are backpropagated via differentiable rasterization.

\subsection{Filtering and Ranking}
To ensure quality, our systematic pipeline selects the best candidates using VLM assessment and quantitative metrics.
% To ensure consistent quality, we establish a systematic evaluation pipeline selecting the most successful progressive illusion sketches from multiple candidates using vision-language model assessment and quantitative metrics.

\begin{figure}[t]
\includegraphics[width=\columnwidth]{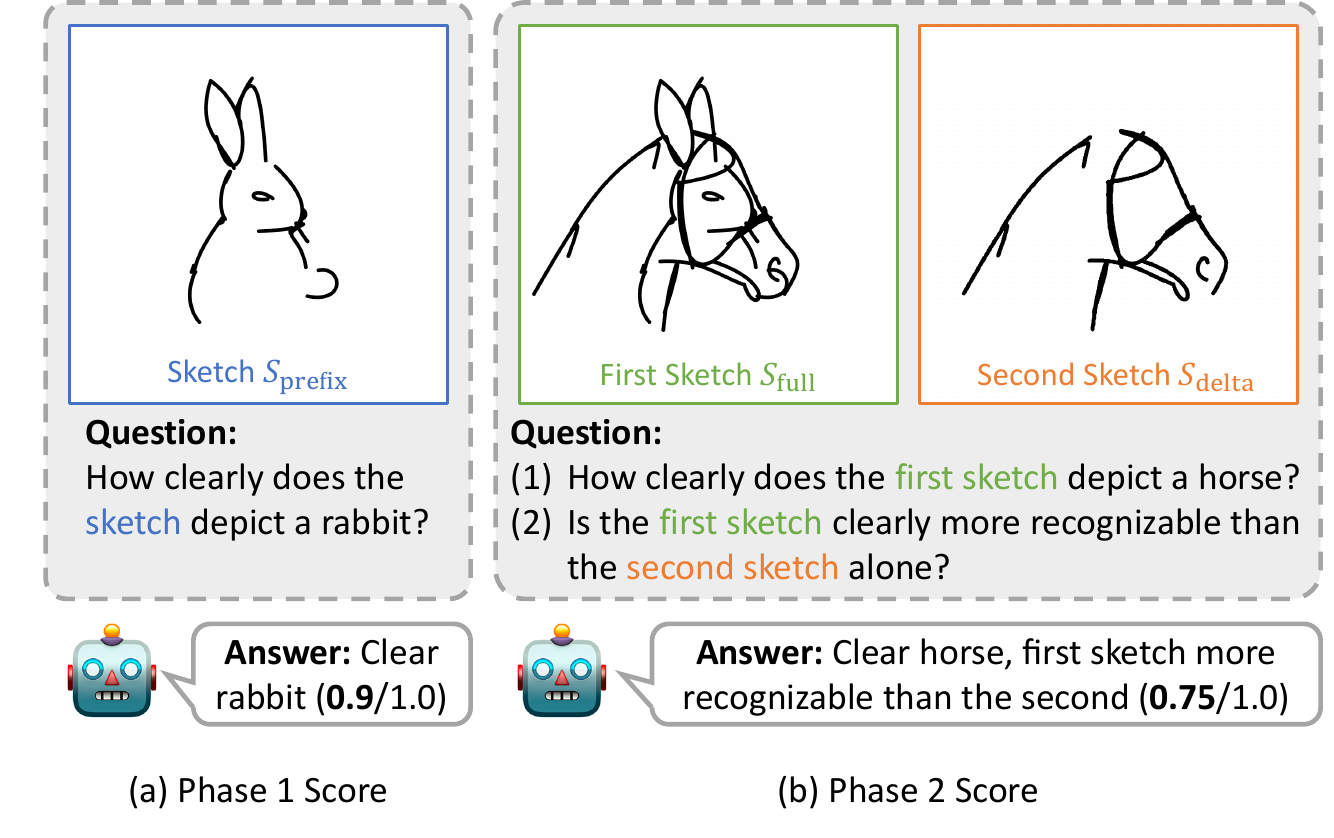}
  \caption{
  \textbf{VLM-based evaluation and ranking pipeline.}
  We employ GPT-4o to assess the quality of illusion sketches.
\textbf{(a) For Phase 1}, the model evaluates the recognizability of the prefix sketch ($S_\text{prefix}$).
\textbf{(b) For Phase 2}, the model evaluates the full sketch ($S_\text{full}$) while simultaneously comparing it against the delta strokes ($S_\text{delta}$). This comparison ensures that the prefix strokes provide \textbf{essential structural scaffolding} for the second concept, rather than being merely overwritten. High scores are awarded only when $S_\text{full}$ is significantly more recognizable than $S_\text{delta}$ alone.
}
  \label{fig:gpt-score-example}
\end{figure}

\paragraph{VLM-based Quality Assessment.} We employ GPT-4o to evaluate four dimensions (Fig. \ref{fig:gpt-score-example}). \emph{Phase recognizability} and \emph{Single-object integrity} ensure semantic accuracy and coherence. \emph{Illusion quality} validates the prefix's structural contribution by confirming $S_{\text{full}}$ is significantly more recognizable than $S_{\text{delta}}$ alone. \emph{Sketch quality} penalizes visual clutter. Each phase receives individual scores across these dimensions, and candidates failing minimum thresholds are filtered.

% \paragraph{VLM-based Quality Assessment.} We employ GPT-4o to evaluate each sketch along four dimensions, as illustrated in Fig. \ref{fig:gpt-score-example}. Phase recognizability measures whether prefix sketch $S_{\text{prefix}}$ and full sketch $S_{\text{full}}$ accurately depict their intended concepts. Single-object integrity verifies each phase presents one coherent object. Illusion quality evaluates structural contribution by comparing $S_{\text{full}}$ against delta strokes $S_{\text{delta}}$ rendered alone (Fig. \ref{fig:gpt-score-example}(b)), requiring $S_{\text{full}}$ demonstrating significantly greater recognizability to confirm prefix strokes participate meaningfully. Sketch quality assesses visual aesthetics, penalizing excessive fill and clutter. We filter candidates failing minimum quality thresholds.

\paragraph{Ranking Strategies.} GPT-based ranking (Fig. \ref{fig:ranking_gpt}) favors semantic accuracy: $\mathcal{R}_{\text{GPT}} = \text{Score}_{\text{Phase 1}} \cdot \text{Score}_{\text{Phase 2}}$. Metric-based ranking (Fig. \ref{fig:ranking_metric}) emphasizes perceptual contrast~\cite{luo2025shadowdraw} by penalizing independent delta stroke quality:
\begin{equation} \small
S_{\text{CLIP}} = (\text{CLIP}_\text{p1} \cdot \text{CLIP}_\text{p2}) / \text{CLIP}_\text{delta}^2,
\end{equation}
\begin{equation} \small
S_{\text{IR}} = \Phi(\text{IR}_\text{p1})^2 + \Phi(\text{IR}_\text{p2})^2 - \Phi(\text{IR}_\text{delta})^2,
\end{equation}
\begin{equation} \small
S_{\text{HPS}} = \text{HPS}_\text{p1}^2 + \text{HPS}_\text{p2}^2 - \text{HPS}_\text{delta}^2,
\end{equation}
where $\Phi(\cdot)$ is the standard Gaussian CDF. The final score $\mathcal{R} = S_{\text{CLIP}} \cdot S_{\text{IR}} \cdot S_{\text{HPS}}$ ensures the prefix provides substantial structural contribution.

% \paragraph{Ranking Strategies.} GPT-based ranking combines phase scores as $\mathcal{R}_{\text{GPT}} = \text{Score}_{\text{Phase 1}} \cdot \text{Score}_{\text{Phase 2}}$,
% favoring high semantic accuracy and recognizability across both phases. Metric-based ranking identifies configurations with strong perceptual contrast following \cite{luo2025shadowdraw}:
% \begin{equation} \small
% S_{\text{CLIP}} = (\text{CLIP}_{p1} \cdot \text{CLIP}_{p2}) / \text{CLIP}_{delta}^2,
% \end{equation}
% \begin{equation} \small
% S_{\text{IR}} = \Phi(\text{IR}_{p1})^2 + \Phi(\text{IR}_{p2})^2 - \Phi(\text{IR}_{delta})^2,
% \end{equation}
% \begin{equation} \small
% S_{\text{HPS}} = \text{HPS}_{p1}^2 + \text{HPS}_{p2}^2 - \text{HPS}_{delta}^2,
% \end{equation}
% where $\Phi(\cdot)$ denotes the CDF of the standard Gaussian. The final metric-based ranking is $\mathcal{R} = S_{\text{CLIP}} \cdot S_{\text{IR}} \cdot S_{\text{HPS}}$.
% By penalizing high alignment of delta strokes alone, this rewards configurations where prefix strokes provide substantial structural contribution, yielding stronger illusion effects. We select top-k candidates based on either strategy.

\begin{figure}[t]
\includegraphics[width=\columnwidth]{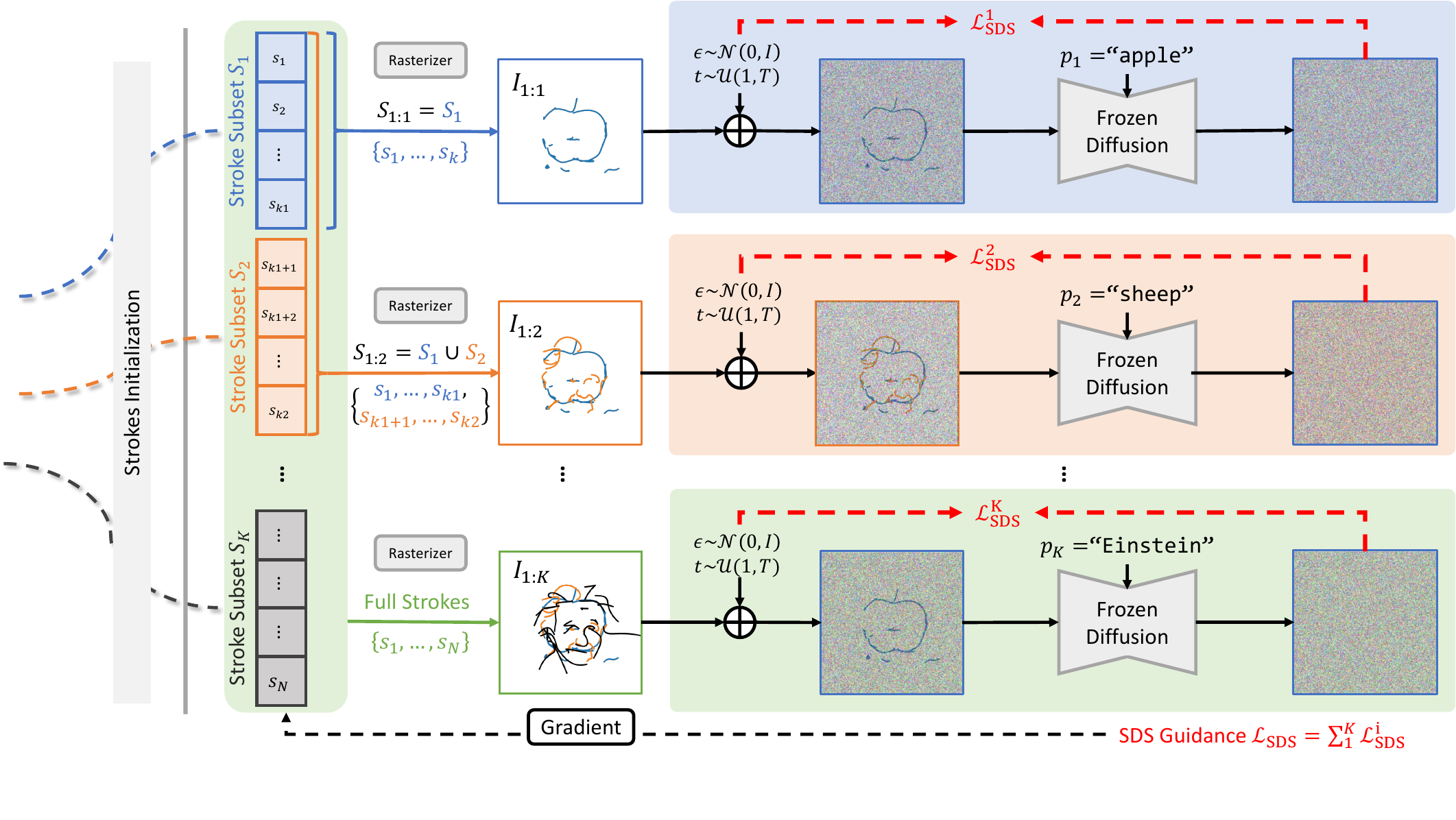}
  \caption{
\textbf{Multi-phase pipeline.} We scale to $K$ phases (e.g., Apple$\to$Sheep$\to$Einstein) using cumulative stroke subsets ($S_1, \ldots, S_K$). Parallel branches optimize each cumulative sketch $I_{1:i}$ against prompt $p_i$. \textbf{Joint optimization} ensures early strokes receive gradients from all subsequent losses ($\sum \mathcal{L}_{\text{SDS}}^i$), creating a structure primed for the entire evolutionary sequence.
  % \textbf{Multi-phase optimization pipeline.} Our framework scales to $K$-phase illusions (e.g., ``Apple'' $\to$ ``Sheep'' $\to$ ``Einstein'') by partitioning strokes into cumulative subsets ($S_1, \ldots, S_K$). We employ parallel branches to render the cumulative sketch $I_{1:i}$ for each phase $i$, conditioned on its specific text prompt $p_i$. Crucially, \textbf{joint optimization} ensures that early stroke subsets (e.g., $S_1$) receive gradient guidance from all subsequent objectives ($\sum \mathcal{L}_{\text{SDS}}^i$), compelling them to form an ``Apple'' that is structurally primed to evolve into a ``Sheep'' and finally an ``Einstein.''
  }
  \label{fig:multiphase}
\end{figure}

\subsection{Extension to Multi-Phase Illusions}
Our framework naturally scales to $K$-phase illusions $(A_1, \ldots, A_K)$ by partitioning strokes into disjoint subsets $S_1, \ldots, S_K$. Each cumulative prefix $S_{1:i} = \bigcup_{j=1}^i S_j$ renders concept $A_i$. We employ parallel branches (Fig.~\ref{fig:multiphase}) to jointly optimize all parameters, rendering $I_{1:i}$ conditioned on prompt $p_i$. This ensures early strokes (e.g., $S_1$) receive gradients from all subsequent branches, coordinating cumulative interpretations. We extend the overlay loss to penalize overlap between $S_{1:i}$ and the next subset $S_{i+1}$:
\begin{equation} \small
\mathcal{L} = \sum_{i=1}^{K} \mathcal{L}_{\text{SDS}}^{i} + \sum_{i=1}^{K-1} \lambda_{\text{overlay}}^{i} \mathcal{L}_{\text{overlay}}^{i}.
\end{equation}

\section{Experiments}

% In this section, we validate our progressive illusion sketch generation framework through comprehensive experiments. We first detail our experimental setup, present results demonstrating our method's superiority, conduct ablation studies, and showcase diverse applications.

\subsection{Experimental Setup} 
% \paragraph{Baseline.}
% We adapt state-of-the-art methods: Nano Banana Pro (raster), SketchAgent~\cite{vinker2025sketchagent}, and SketchDreamer~\cite{qu2023sketchdreamer} (vector). We design two protocols:
% (1) \textbf{Text-to-illusion}: Baselines generate sketches sequentially (prefix from $p_1$, then full from $p_2$). For the raster-based Nano Banana Pro, we enforce the progressive constraint by overlaying the prefix onto the final output, whereas vector baselines natively support stroke addition.
% (2) \textbf{Ours-to-illusion}: We provide our optimized prefix sketches as input to evaluate whether baselines can complete the transformation given an ideal structural foundation.

\paragraph{Baseline.}
% We adapt state-of-the-art methods: Nano Banana Pro (raster), SketchAgent~\cite{vinker2025sketchagent}, and SketchDreamer~\cite{qu2023sketchdreamer} (vector), \huaish{as well as two image-based methods, CLIPasso~\cite{vinker2022clipasso} and ControlSketch~\cite{arar2025swiftsketch}, to the progressive illusion task.} We design two evaluation protocols:
% \huaish{(1) \textbf{Text-to-illusion}: Baselines generate sketches sequentially, producing the prefix from $p_1$ and the full sketch from $p_2$. Nano Banana Pro enforces the progressive constraint via prefix overlaying, while vector baselines natively support stroke addition. CLIPasso and ControlSketch require image input; we supply SDXL-generated~\cite{podell2023sdxlimprovinglatentdiffusion} reference images from the same text prompts, with stroke count and segment number matched to our setting. SketchDreamer is evaluated at its default of 5-segment cubic B\'ezier curves, as reducing to our 1-segment setting causes its reinitialization mechanism to discard short curves and substantially degrade output quality.}
% (2) \textbf{Ours-to-illusion}: We provide our optimized prefix sketches as input to evaluate whether baselines can complete the transformation given an ideal structural foundation.
We adapt state-of-the-art methods to the progressive illusion task: Nano Banana Pro (raster), SketchAgent~\cite{vinker2025sketchagent}, SketchDreamer~\cite{qu2023sketchdreamer} (vector), and two image-based methods, CLIPasso~\cite{vinker2022clipasso} and ControlSketch~\cite{arar2025swiftsketch}. We design two protocols: (1) \textbf{Text-to-illusion}: Baselines generate sketches sequentially---prefix from $p_1$, full sketch from $p_2$. Nano Banana Pro enforces the progressive constraint via prefix overlaying; vector baselines natively support stroke addition. CLIPasso and ControlSketch require image input, so we supply SDXL-generated~\cite{podell2023sdxlimprovinglatentdiffusion} references from the same prompts, with matched stroke count and segments. SketchDreamer uses its default 5-segment cubic Bézier curves, as our 1-segment setting triggers reinitialization that degrades quality. (2) \textbf{Ours-to-illusion}: We supply our optimized prefix sketches to test whether baselines can complete the transformation given an ideal structural foundation.

\paragraph{Data.}
Our evaluation dataset comprises 64 common objects spanning diverse categories. We randomly sample pairs to form $(p_1, p_2)$ combinations, run multiple optimization iterations per pair, then apply filtering and ranking to select top-k results for evaluation.

\paragraph{Implementation Details}
We implement our framework using Stable Diffusion v1.5 for Score Distillation Sampling guidance on an NVIDIA RTX 4090 GPU. We optimize stroke parameters $\theta$ for 2,000 iterations using Adam optimizer with guidance scale 100 and overlay loss weight $\lambda_{\text{overlay}} = 0.1$. Generation requires approximately 13 minutes for two-phase and 15 minutes for three-phase illusions.

\paragraph{Metrics.}
% For quantitative evaluation, we employ both standard and specialized metrics to assess illusion quality. We employ CLIP score computed as the minimum across all phases to measure semantic alignment. Beyond standard metrics, we define two illusion-specific measures. Structural concealment evaluates whether prefix strokes $S_{\text{prefix}}$ contribute substantively to the full sketch rather than being occluded by delta strokes $S_{\text{delta}}$. For any metric $M \in \{\text{CLIP}, \text{ImageReward}, \text{HPS}\}$, where CLIP \cite{hessel2022clipscorereferencefreeevaluationmetric}, ImageReward \cite{xu2023imagereward}, and HPS \cite{wu2023human} are established image quality metrics, we compute inspired by \cite{luo2025shadowdraw}:
% \begin{equation}
% C^M_{\text{struct}} = M_{\text{full}} - M_{\text{delta}}.
% \end{equation}
% Higher scores indicate prefix strokes retain significant structural roles. Semantic concealment measures whether non-current phase semantics are effectively hidden. Following \cite{geng2024visualanagrams}, we compute:
% \begin{equation}
% C_{\text{semantic}} = \text{tr}(\text{softmax}(S/\tau)),
% \end{equation}
% where $S$ is the CLIP image-text similarity matrix and $\tau$ is temperature. Higher scores indicate clear phase-specific semantics, implying successful concealment of other phases.
For quantitative evaluation, we employ both standard and specialized metrics to assess illusion quality. We use CLIP score computed as the minimum across all phases to measure semantic alignment. Beyond standard metrics, we define two illusion-specific measures. Structural concealment evaluates whether prefix strokes contribute substantively to the full sketch rather than being occluded by delta strokes. For any metric $M \in \{\text{CLIP}, \text{ImageReward}, \text{HPS}\}$ \cite{hessel2022clipscorereferencefreeevaluationmetric, xu2023imagereward, wu2023human}, we compute: $C^M_{\text{struct}} = M_{\text{full}} - M_{\text{delta}}.$ Higher scores indicate prefix strokes retain significant structural roles. Semantic concealment measures whether non-current phase semantics are effectively hidden. Following \cite{geng2024visualanagrams}, we compute: 
\begin{equation} \small
C_{\text{semantic}} = \text{tr}(\text{softmax}(S/\tau)),
\end{equation}
where $S$ is the CLIP image-text similarity matrix and $\tau$ is temperature. Higher scores indicate clear phase-specific semantics.

% we adopt their CLIP-based trace metric over the image-text similarity matrix, where higher scores indicate clear phase-specific semantics.

We further conduct two user studies with 143 participants for additional quantitative validation. The first compares our top-1 result against baselines across five prompt pairs. The second assesses our ranking pipeline by asking participants to select satisfactory results from our top-4 outputs across four prompt pairs, evaluating both technical performance and practical user satisfaction.

\begin{table}[t]
\centering
\small
\caption{
\textbf{Quantitative comparison.} (a) Vector baselines lack quality; Nano Banana fails coverage ($\sim$35\%). (b) Extending our Phase 1 helps, but still lags behind (c), validating joint optimization. (c) Ours achieves top metrics with 100\% coverage. Best in \textbf{bold}, second \underline{underlined}.
}
% \textbf{Quantitative comparison.} We report CLIP, Concealment, and Coverage across: (a) standard baselines, (b) baselines extending our Phase 1, and (c) our full pipeline.
% \textbf{(1) Baseline Limitations}: In (a), vector methods lack semantic quality, while Nano Banana Pro fails coverage ($\sim$35\%) due to destructive editing.
% \textbf{(2) Joint vs. Sequential}: In (b), baselines improve using our structural priors but still underperform compared to (c), validating the superiority of joint optimization over sequential extension.
% \textbf{(3) Performance}: Our method (c) achieves the best semantic and illusion scores with strict \textbf{100\% coverage}. Best in \textbf{bold}, second \underline{underlined}.
\resizebox{\columnwidth}{!}{%
\begin{tabular}{cl|c|c|ccc|c|c}
\toprule
& & Phase 1 & CLIP $\uparrow$ & \multicolumn{3}{c|}{Concealment (structural)} & $C_{\text{semantic}}$  & Coverage \\
\cmidrule(lr){5-7}
& Method &  Source & Avg min & CLIP $\uparrow$ & IR $\uparrow$ & HPS $\uparrow$ & CLIP $\uparrow$ &  (\%) $\uparrow$  \\
\midrule
& CLIPasso        & -  & \textbf{32.213} & \underline{1.690} & 0.090 & 0.004 & \textbf{1.000} & \textbf{100.0\%} \\
& ControlSketch       & -  & 27.524 & -2.378 & -0.789 & -0.018 & 0.875 & \textbf{100.0\%} \\
(a) & SketchDreamer      & -      & 24.803 & -0.393 & 0.338 & 0.011 & 0.887 & \textbf{100.0\%} \\
& SketchAgent        & -        & 24.393 & -2.544 & 0.095 & 0.000 & 0.752 & \textbf{100.0\%} \\
& Nano Banana Pro    & -    & 26.821 & -2.774 & -0.663 & -0.019 & 0.875 & 34.9\% \\
\midrule
& SketchDreamer      & Ours & 28.148 & 0.060 & 0.302 & 0.011 & 0.961 & \textbf{100.0\%} \\
(b) & SketchAgent        & Ours & 24.019 & -2.778 & 0.080 & 0.003 & 0.762 & \textbf{100.0\%} \\
& Nano Banana Pro    & Ours & 28.903 & -1.065 & -0.426 & -0.014 & 0.958 & 35.2\% \\
\midrule
\multirow{2}{*}{(c)} & Ours (GPT-ranking)    & -- & 29.873 & 1.668 & \underline{0.839} & \underline{0.023} & \underline{0.983} & \textbf{100.0\%} \\
& Ours (Metric-ranking) & -- & \underline{30.044} & \textbf{3.282} & \textbf{1.237} & \textbf{0.029} & 0.980 & \textbf{100.0\%} \\
\bottomrule
\end{tabular}
}
\label{tab:text_to_illusion}
\end{table}

\begin{figure*}[t]
  \centering
  \includegraphics[width=\textwidth]{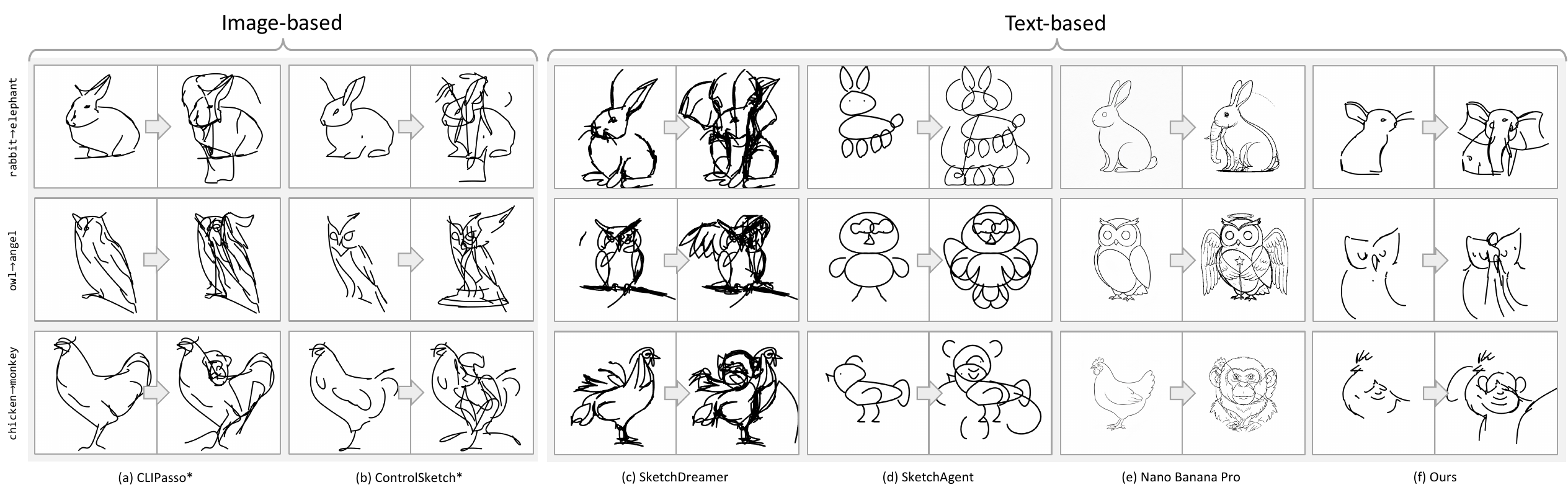}
    \caption{
    \textbf{Qualitative comparisons.} 
    We compare against CLIPasso~\cite{vinker2022clipasso} and ControlSketch~\cite{arar2025swiftsketch} (image-based, denoted $^*$), SketchDreamer~\cite{qu2023sketchdreamer}, SketchAgent~\cite{vinker2025sketchagent}, and Nano Banana Pro (text-based). Image-based methods use SDXL-generated~\cite{podell2023sdxlimprovinglatentdiffusion} references from the same prompts.
    (a, b) CLIPasso and ControlSketch follow reference contours too closely, failing to integrate prefix with delta strokes.
    (c) SketchDreamer produces noisy strokes with severe clutter.
    (d) SketchAgent yields overly abstract results with low recognizability.
    (e) Nano Banana Pro relies on \textbf{destructive editing}, violating the progressive constraint despite high image quality.
    (f) Ours generates clean, structurally consistent sketches where prefix strokes are creatively repurposed (e.g., rabbit ears becoming elephant ears).
    Additional results and optimization visualizations are in the supplement.}
  \label{fig:text-to-illusion}
\end{figure*}

\begin{figure*}[t]
  \includegraphics[width=\textwidth]{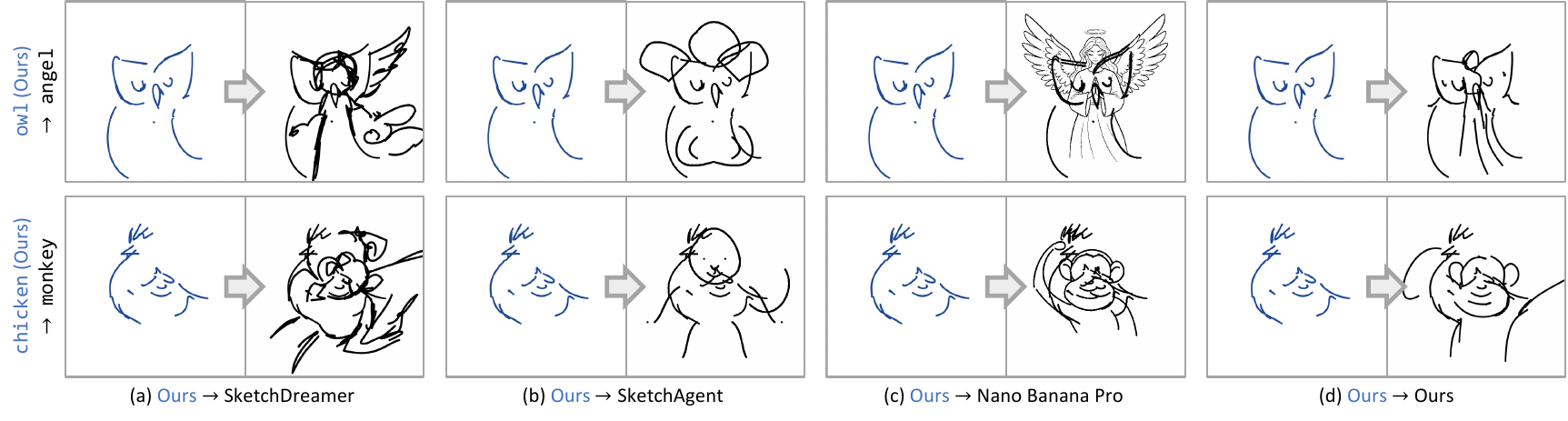}
  \caption{
  \textbf{Phase 2 extension with fixed prefix (ours).}
  We evaluate how methods extend a fixed Phase 1 sketch generated by our method.
Interestingly, baselines produce better Phase 2 results here than in Fig.~\ref{fig:text-to-illusion} (where they generate Phase 1 themselves). This indicates that our Phase 1 strokes inherently embed structural cues for the second concept, validating that our joint optimization successfully finds a versatile common subspace.
However, comparing (a-c) with (d), our method still achieves the highest success rate and structural consistency, as $S_\text{delta}$ is jointly optimized with the prefix rather than sequentially appended.
  }
  \label{fig:ours-to-illusion}
\end{figure*}

\subsection{Results and Analysis} 
As shown in Tab.~\ref{tab:text_to_illusion}(a,c), our method substantially outperforms baselines in CLIP and concealment scores, achieving 100\% coverage versus Nano Banana Pro's 34.9\%. Among image-based baselines, CLIPasso attains the highest Phase-1 CLIP score via direct image conditioning, yet concealment remains low; ControlSketch shows negative concealment scores. Both optimize strokes to reproduce the input image, so prefix strokes lack flexibility for a second semantic interpretation. Fig.~\ref{fig:text-to-illusion} highlights characteristic failures: clutter (SketchDreamer), oversimplification (SketchAgent), destructive editing (Nano Banana Pro), and prefix strokes too tightly bound to references for recontextualization (CLIPasso, ControlSketch). Tab.~\ref{tab:text_to_illusion}(b,c) and Fig.~\ref{fig:ours-to-illusion} show that with fixed prefixes, baselines improve, suggesting our prefixes embed implicit structural cues (``common subspace''), yet remain substantially inferior to ours, confirming that joint optimization is essential for seamless integration.

\begin{figure}[t]
\centering
\small
\resizebox{\columnwidth}{!}{
\begin{tabular}{cc}
\includegraphics[height=0.3\columnwidth]{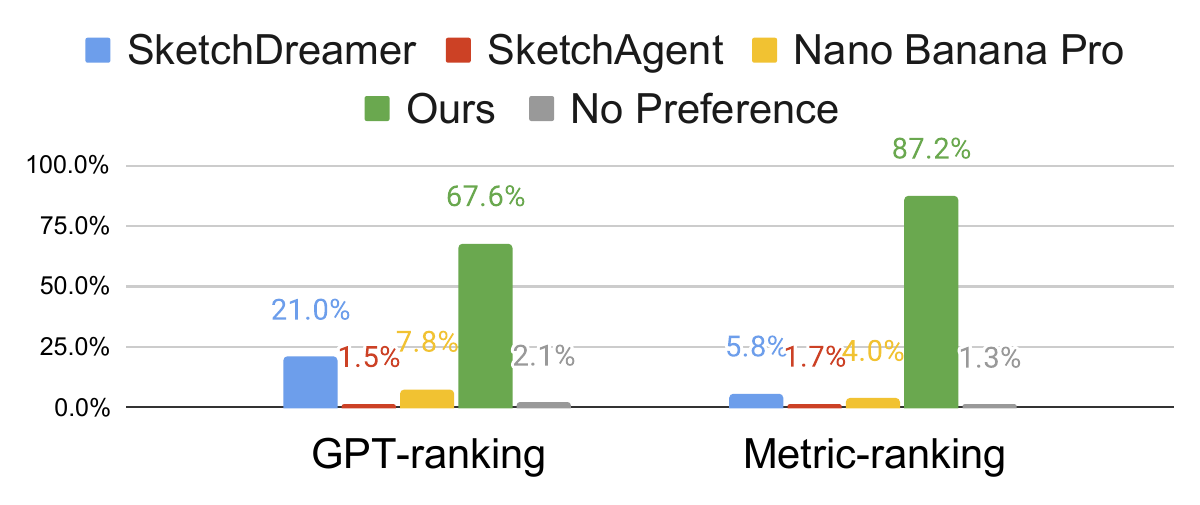} & 
\includegraphics[height=0.3\columnwidth]{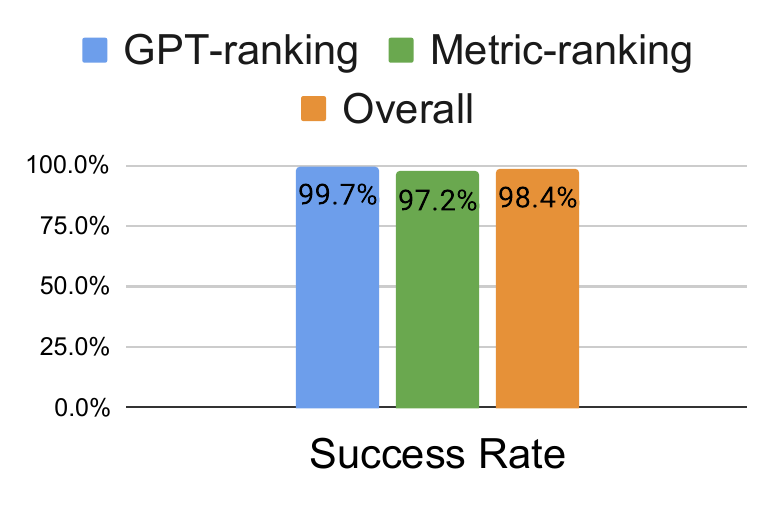} \\
% (a) Comparison & (b) Success Rate \\
\end{tabular}
}
\caption{
\textbf{User study.} \textbf{(Left) Preference}: Participants overwhelmingly favor our method (\textcolor{mygreen}{\textbf{green}}) over baselines across both ranking strategies. \textbf{(Right) Reliability}: A high success rate (>97\%) confirms that our pipeline consistently yields valid illusions, ensuring robustness against the inherent stochasticity of the generation process.
% \textbf{User study results.} \textbf{(\emph{Left}) Human preference}: Participants overwhelmingly favor our method (\textcolor{mygreen}{\textbf{green}}) over baselines regardless of the ranking strategy used (GPT vs. Metric), confirming our superior illusion quality. \textbf{(\emph{Right}) Pipeline reliability}: The high success rate (>97\%) of top candidates validates that our automated filtering and ranking framework aligns closely with human perception in selecting valid illusion sketches.
% \textbf{User study.}
% (a) 不論是 GPT-ranking 或是 Metric-ranking，我們的方法都 achieves the highest user prefernce ratio，都明顯優於所有 baseline。顯示我們的方法生成的結果更符合人類對於好的 illusion sketch 的判斷與 quantitative 一致
% (b) 從 Success Rate 圖表，顯示 GPT-ranking 和 Metric-ranking 皆達到 97\% 以上的成功率，說明我們的 ranking and filtering framework 是有效的
}
\label{fig:userstudy}
\end{figure}

\paragraph{User Studies.}
Our user studies strongly reinforce these findings. In comparisons against baselines, participants selected our method in 67.7\% of GPT-ranking and 87.1\% of Metric-ranking cases (Fig. \ref{fig:userstudy}(a)). Our ranking pipeline demonstrates strong reliability with over 98\% overall satisfaction rates (Fig. \ref{fig:userstudy}(b)), thoroughly validating our framework's effectiveness.

\subsection{Ablation Studies}

\begin{figure}[t]
  \centering
  \includegraphics[width=\columnwidth]{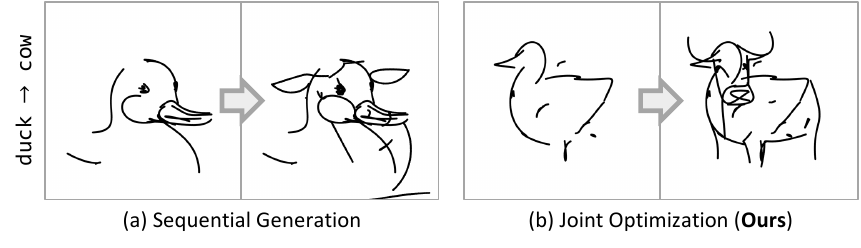}
  \caption{
  \textbf{Ablation on optimization strategy.}
\textbf{(a) Sequential generation} yields a rigid Phase 1, creating structural conflicts (e.g., the duck's beak) that fail Phase 2 repurposing.
\textbf{(b) Joint optimization (Ours)} identifies a \textbf{common structural subspace}, yielding a versatile Phase 1 where features serve both interpretations (e.g., the beak doubles as the cow's ear).
%   \textbf{Ablation on optimization strategy.}
% \textbf{(a) Sequential generation} (greedy) produces a rigid Phase 1. Specific features (e.g., the duck's beak) create structural conflicts that prevent stroke repurposing for Phase 2, resulting in failure.
% \textbf{(b) Joint optimization (Ours)} optimizes both phases simultaneously to identify a \textbf{common structural subspace}. This yields a versatile Phase 1 compatible with the Phase 2 illusion (e.g., the beak doubles as the cow's ear).
  }
  \label{fig:ablation-optimization}
\end{figure}

\paragraph{Optimization Strategy}
We evaluate our joint optimization approach against a sequential alternative that first optimizes prefix strokes independently for the initial concept, then fixes these parameters and optimizes only delta strokes. As shown in Fig. \ref{fig:ablation-optimization}(a), this sequential approach produces rigid prefix structures where specific features conflict with the final object, resulting in failed illusion transitions. The prefix optimization focuses solely on the initial concept without considering final target requirements. In contrast, our joint optimization (Fig. \ref{fig:ablation-optimization}(b)) updates both stroke sets simultaneously, enabling continuous coordination. This allows the framework to discover a common structural subspace where prefix strokes both represent the initial concept and integrate naturally into the final representation. The results demonstrate improved visual consistency and smooth transitions, confirming that joint optimization is essential for high-quality progressive illusion sketches.

\begin{figure}[t]
  \centering
  \includegraphics[width=\columnwidth]{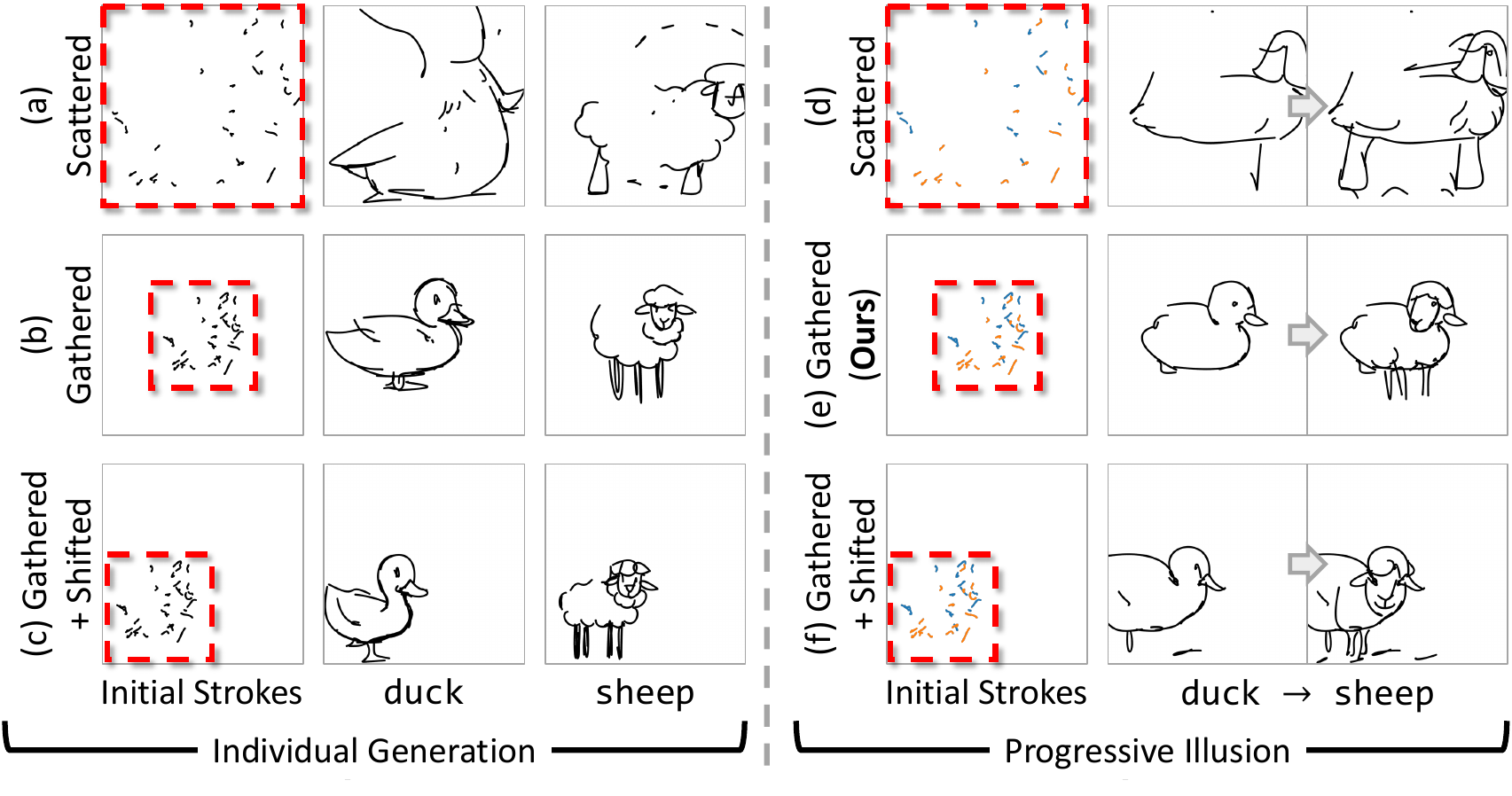}
  \caption{
  \textbf{Ablation on stroke initialization.}
\textbf{(a, d) Scattered} fails to aggregate strokes, resulting in disconnected artifacts.
\textbf{(c, f) Shifted} yields valid sketches, proving that \textbf{spatial concentration} is critical for convergence, though it risks boundary cropping.
\textbf{(b, e) Centered (Ours)} offers the optimal balance, ensuring structural integrity without clipping.
%   \textbf{Ablation on stroke initialization.}
% \textbf{(a, d) Scattered initialization} fails to aggregate widely spaced strokes, resulting in disconnected artifacts.
% \textbf{(c, f) Shifted gathered} yields valid sketches, proving that \textbf{spatial concentration} is critical for convergence, though it risks boundary cropping.
% \textbf{(b, e) Centered gathered (Ours)} offers the optimal balance, ensuring structural integrity and efficient convergence without clipping.
  }
  \label{fig:ablation-init}
\end{figure}

\paragraph{Stroke Initialization.}
% Since our objective function is highly non-convex, initialization is critical for convergence. \huaish{Centered initialization is standard practice in SDS-based sketch methods~\cite{jain2023vectorfusion, xing2023diffsketcher, vinker2022clipasso}; in our dual-constraint setting, however, it is especially necessary since prefix strokes must form a spatially coherent structure valid for both semantic interpretations simultaneously.} Fig.~\ref{fig:ablation-init} shows that spatial concentration is paramount; scattered initialization fails to capture essential semantic features. In contrast, both centered and shifted gathered configurations succeed, indicating that local stroke density outweighs absolute position. We therefore adopt centered gathered initialization to balance density with spatial coverage, avoiding potential boundary clipping. \huaish{Quantitatively, scattered initialization scores only 3.759 on concealment CLIP compared to 5.723 for centered gathered initialization (Tab.~\ref{tab:ablation}, Abl.~1 vs.~Ours), validating that spatial concentration is indispensable for our dual-constraint task.}
Since our objective is highly non-convex, initialization is 
critical for convergence. Centered initialization is 
standard in SDS-based sketch 
methods~\cite{jain2023vectorfusion,xing2023diffsketcher,
vinker2022clipasso}; in our dual-constraint setting it is 
especially necessary, as prefix strokes must form a 
spatially coherent structure valid for both semantic 
interpretations. Fig.~\ref{fig:ablation-init} shows 
spatial concentration is paramount: scattered 
initialization fails to capture essential features, while 
both centered and shifted gathered configurations succeed, 
indicating local stroke density outweighs absolute 
position. We adopt centered gathered initialization to 
balance density with coverage and avoid boundary clipping. 
% \huaish{Quantitatively, scattered initialization scores only 3.759 
% on concealment CLIP versus 5.723 for centered gathered 
% (Tab.~\ref{tab:ablation}, Abl.~1 vs.~Ours), confirming 
% spatial concentration is indispensable.}

% We investigate the impact of stroke initialization strategies on optimization convergence and sketch quality. Since our objective function is highly non-convex, initialization critically influences convergence to favorable local minima. Fig. \ref{fig:ablation-init} shows that spatial concentration fundamentally determines the success of optimization. Scattered initialization fails to produce coherent results, yielding incomplete sketches that lack essential semantic features even for individual generation tasks. In contrast, both centered gathered and shifted gathered configurations successfully generate high-quality illusion sketches, indicating that positional translation has minimal impact when adequate stroke density is preserved. This reveals that local stroke concentration enables the optimization to capture fine-grained details necessary for recognizable representations. We adopt centered gathered initialization as our default strategy, balancing stroke density with spatial coverage while avoiding potential boundary issues observed with shifted configurations.

\begin{figure}[t]
  \centering
  \includegraphics[width=\columnwidth]{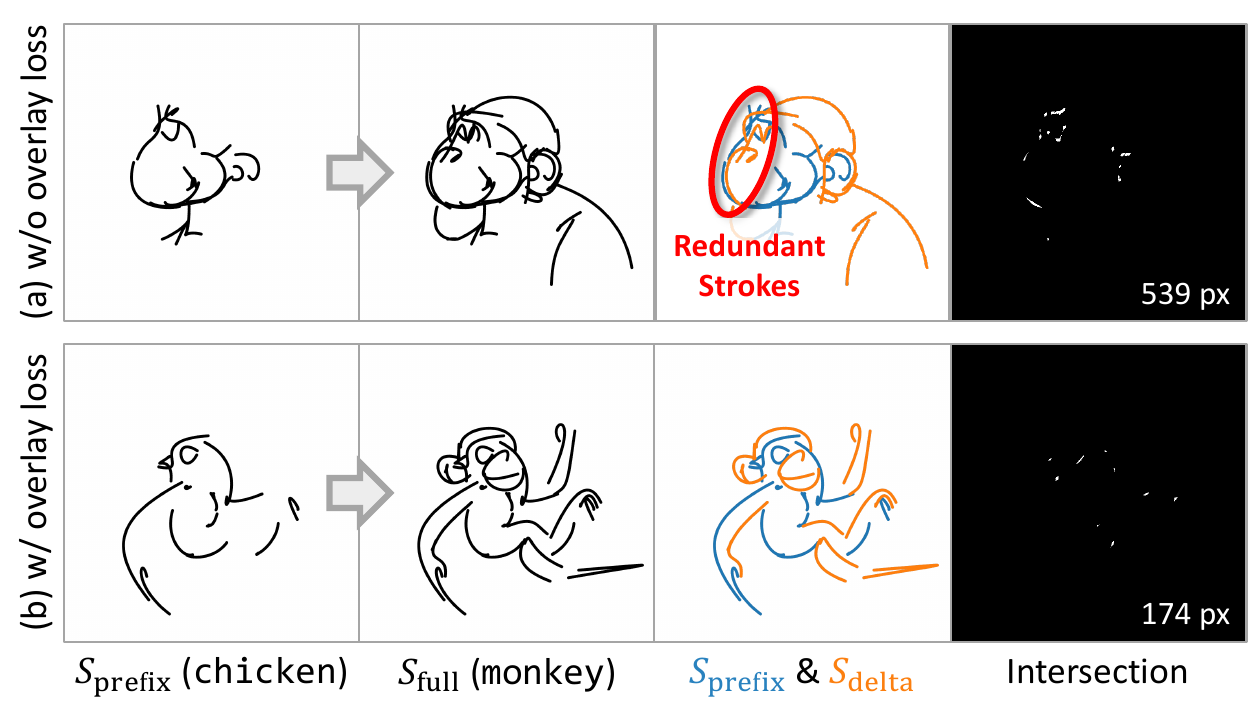}
  \caption{
  \textbf{Ablation of overlay loss ($\mathcal{L}_\text{overlay}$).}
(a) Without $\mathcal{L}_\text{overlay}$, the model generates redundant strokes atop existing ones to satisfy the semantic target, resulting in visual clutter (\textcolor{red}{red circle}) and high intersection artifacts.
(b) With $\mathcal{L}_\text{overlay}$, the generated strokes (\textcolor{myorange}{$S_\text{delta}$}) become spatially complementary to the prefix (\textcolor{myblue}{$S_\text{prefix}$}), avoiding collisions to produce a clean, coherent line drawing.
%   \textbf{Ablation study of the overlay Loss ($\mathcal{L}_\text{overlay}$).}
%   We visualize the impact of penalizing stroke overlaps.
% (a) Without $\mathcal{L}_\text{overlay}$, the model tends to blindly generate new strokes on top of existing ones to satisfy the semantic target, resulting in visual clutter (\textcolor{red}{red circle}) and heavy artifacts (high intersection pixel count).
% (b) With our overlay loss, the generated strokes (\textcolor{myorange}{$S_\text{delta}$}) are encouraged to be spatially complementary to the prefix (\textcolor{myblue}{$S_\text{prefix}$}), avoiding collisions and producing a clean, coherent line drawing.
  }
  \label{fig:ablation-overlay}
\end{figure}

\paragraph{Overlay Loss.} 
We validate the necessity of $\mathcal{L}_{\text{overlay}}$. 
As shown in Fig.~\ref{fig:ablation-overlay}(a), without it, 
semantic guidance alone fails to prevent spatial redundancy, 
producing delta strokes that clutter the prefix; 
% \huaish{quantitatively, removing $\mathcal{L}_{\text{overlay}}$ 
% drops concealment CLIP from 5.723 to 2.421 
% (Tab.~\ref{tab:ablation}, Abl.~3 vs.~Ours).} 
$\mathcal{L}_{\text{overlay}}$ penalizes overlap, enforces 
spatial complementarity, and substantially reduces 
intersection artifacts 
(Fig.~\ref{fig:ablation-overlay}(b)). Crucially, it 
promotes structural coherence: prefix strokes integrate 
naturally into the subsequent concept rather than being 
obscured, confirming that geometric constraints are 
essential for clean progressive illusions.

\begin{figure}
  \includegraphics[width=\columnwidth]{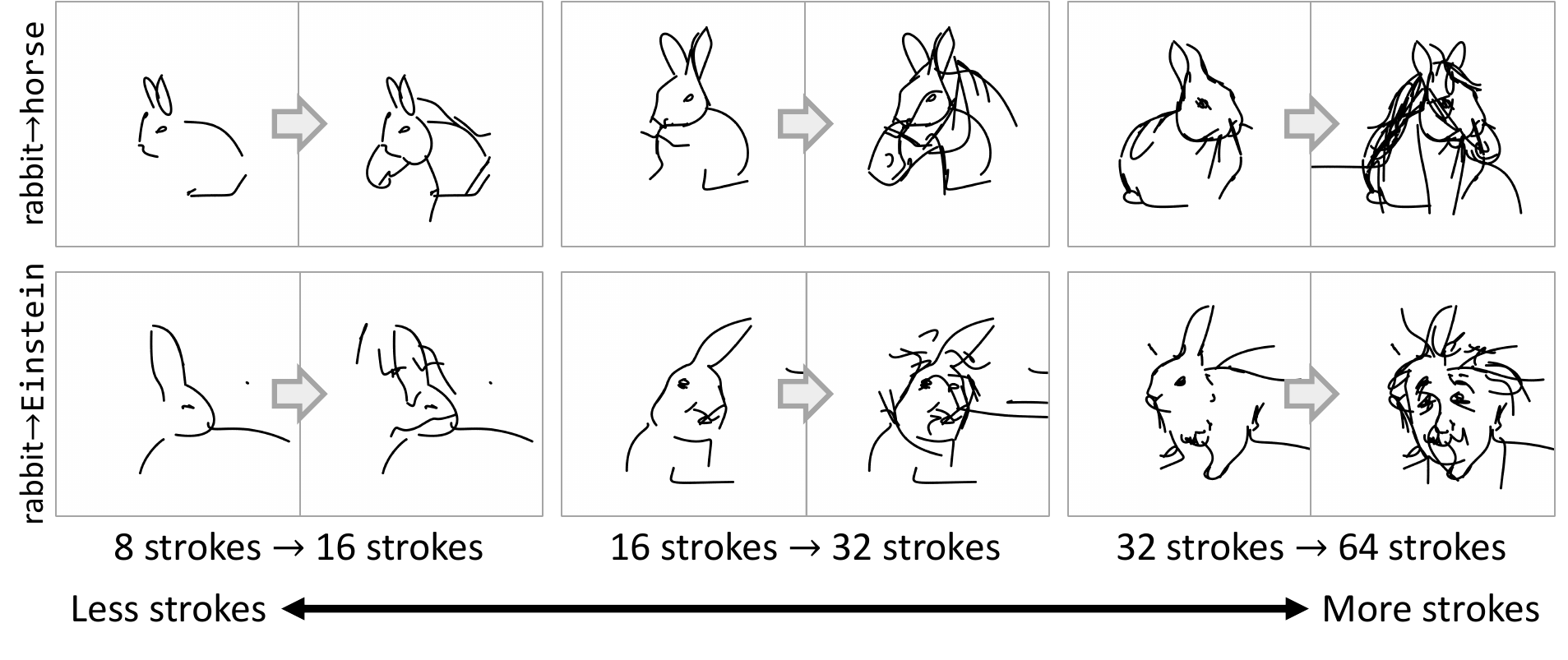}
  \caption{
  \textbf{Analysis of stroke count.} (\emph{Top}) Simple concepts (\texttt{horse}) form recognizable silhouettes with minimal strokes (8$\rightarrow$16). (\emph{Bottom}) While complex concepts (\texttt{Einstein}) require a larger budget (32$\rightarrow$64) to capture essential details. Fewer strokes result in abstraction. Our default (16$\rightarrow$32) balances structural simplicity and semantic fidelity.
% \textbf{Analysis of stroke count.}
% (\emph{Top}) Simple shape-based concepts (\texttt{horse}) form recognizable silhouettes with minimal strokes (8$\rightarrow$16).
% (\emph{Bottom}) Complex concepts (\texttt{Einstein}) require a larger budget (32$\rightarrow$64) to capture essential details. Fewer strokes result in abstraction.
% Our default (16$\rightarrow$32) robustly balances structural simplicity with semantic fidelity across varying levels of complexity.
  }
  \label{fig:ablation-stroke-count}
\end{figure}

\paragraph{Stroke Count.} 
Optimal stroke budget depends on concept complexity (Fig.~\ref{fig:ablation-stroke-count}). Simple transformations (e.g., rabbit-to-horse) succeed with minimal strokes (8--16), whereas complex subjects like Einstein require 32--64 strokes to capture essential details; insufficient budgets compromise recognizability. We therefore adopt a default of 16 prefix strokes and 32 total strokes, robustly balancing structural simplicity with semantic fidelity.

% We investigate how stroke budget affects sketch quality and recognizability across varying concept complexity. Fig. \ref{fig:ablation-stroke-count} demonstrates that optimal stroke count depends critically on semantic and structural complexity of target concepts. Simple shape-based concepts such as the rabbit-to-horse transformation form recognizable silhouettes with minimal strokes, requiring only 8 to 16 strokes to convey clear semantic meaning. Complex concepts such as Einstein require larger budgets of 32 to 64 strokes to capture essential facial features and characteristic details necessary for confident recognition. Insufficient stroke budgets result in abstraction that compromises recognizability. Based on these observations, we adopt a default configuration of 16 prefix strokes and 32 total strokes, which robustly balances structural simplicity with semantic fidelity across varying levels of complexity.

\subsection{Applications}
\paragraph{Technical Versatility.}
We demonstrate versatility beyond standard two-phase scenarios. Fig. \ref{fig:appendix_2_phase} confirms robustness across diverse concept pairs, ranging from structurally similar to semantically distant. Fig. \ref{fig:appendix_3_phase} extends this to three-phase illusions (e.g., apple-to-rabbit-to-pig), showcasing effective multi-target coordination. Furthermore, our framework generalizes to alternative representations, including B-spline curves (Fig. \ref{fig:appendix_bspline}), vector graphics (Fig. \ref{fig:appendix_vector_graph}), and colored sketches (Fig. \ref{fig:appendix_color}), validating the broad applicability of our joint optimization principle.

\paragraph{Practical Applications.}
\textbf{Creative education:} Progressive illusions serve as spatial reasoning exercises, fostering Gestalt perception.
\textbf{Brand and logo design:} Animated illusions bridge two brand identities in a single vector asset for mergers or motion graphics.
\textbf{Physical media steganography:} Outputs support thermochromic printing---a mug shows one concept at rest; heat reveals the transformation.
\textbf{Dynamic visual storytelling:} Native vector output enables arbitrary-resolution rendering and frame-by-frame animation for interactive media.
\textbf{Cognitive science:} The framework generates controlled stimuli for studying temporal semantic perception with calibrated structural overlap.

\section{Conclusion}
We present Stroke of Surprise, the first framework for progressive semantic illusions in vector sketching. By shifting from spatial to temporal dimensions, we enable real-time semantic re-contextualization. Our joint optimization strategy demonstrates that prefix strokes must be "primed" for future semantics. Greedy baselines do not have this ability. Meanwhile, the Overlay Loss ensures structural integration without obfuscation. Evaluations confirm our results are both semantically accurate and perceptually surprising.

\paragraph{Limitations.} Our method inherits limitations from pre-trained diffusion priors; weak SDS guidance for complex structures (e.g., ``scissors'') causes optimization failure. We provide visual examples in the supplementary material.

\begin{acks}
This research was funded by the National Science and Technology Council, Taiwan, under Grants NSTC 112-2222-E-A49-004-MY2 and 113-2628-E-A49-023-. The authors are grateful to Google, NVIDIA, and MediaTek Inc. for their generous donations. Yu-Lun Liu acknowledges the Yushan Young Fellow Program by the MOE in Taiwan.
\end{acks}

%%
%% The next two lines define the bibliography style to be used, and
%% the bibliography file.
\bibliographystyle{ACM-Reference-Format}
\bibliography{sample-base}

%%% -*-BibTeX-*-
%%% Do NOT edit. File created by BibTeX with style
%%% ACM-Reference-Format-Journals [18-Jan-2012].

\begin{thebibliography}{63}

%%% ====================================================================
%%% NOTE TO THE USER: you can override these defaults by providing
%%% customized versions of any of these macros before the \bibliography
%%% command.  Each of them MUST provide its own final punctuation,
%%% except for \shownote{} and \showURL{}.  The latter two
%%% do not use final punctuation, in order to avoid confusing it with
%%% the Web address.
%%%
%%% To suppress output of a particular field, define its macro to expand
%%% to an empty string, or better, \unskip, like this:
%%%
%%% \newcommand{\showURL}[1]{\unskip}   % LaTeX syntax
%%%
%%% \def \showURL #1{\unskip}           % plain TeX syntax
%%%
%%% ====================================================================

\ifx \showCODEN    \undefined \def \showCODEN     #1{\unskip}     \fi
\ifx \showISBNx    \undefined \def \showISBNx     #1{\unskip}     \fi
\ifx \showISBNxiii \undefined \def \showISBNxiii  #1{\unskip}     \fi
\ifx \showISSN     \undefined \def \showISSN      #1{\unskip}     \fi
\ifx \showLCCN     \undefined \def \showLCCN      #1{\unskip}     \fi
\ifx \shownote     \undefined \def \shownote      #1{#1}          \fi
\ifx \showarticletitle \undefined \def \showarticletitle #1{#1}   \fi
\ifx \showURL      \undefined \def \showURL       {\relax}        \fi
% The following commands are used for tagged output and should be
% invisible to TeX
\providecommand\bibfield[2]{#2}
\providecommand\bibinfo[2]{#2}
\providecommand\natexlab[1]{#1}
\providecommand\showeprint[2][]{arXiv:#2}

\bibitem[Arar et~al\mbox{.}(2025)]%
        {arar2025swiftsketch}
\bibfield{author}{\bibinfo{person}{Ellie Arar}, \bibinfo{person}{Yarden Frenkel}, \bibinfo{person}{Daniel Cohen-Or}, \bibinfo{person}{Ariel Shamir}, {and} \bibinfo{person}{Yael Vinker}.} \bibinfo{year}{2025}\natexlab{}.
\newblock \showarticletitle{Swiftsketch: A diffusion model for image-to-vector sketch generation}. In \bibinfo{booktitle}{\emph{Proceedings of the Special Interest Group on Computer Graphics and Interactive Techniques Conference Conference Papers}}. \bibinfo{pages}{1--12}.
\newblock


\bibitem[B{\'e}zier(1968)]%
        {bezier1968}
\bibfield{author}{\bibinfo{person}{Pierre~E. B{\'e}zier}.} \bibinfo{year}{1968}\natexlab{}.
\newblock \bibinfo{booktitle}{\emph{How {Renault} Uses Numerical Control for Car Body Design and Tooling}}.
\newblock \bibinfo{type}{{T}echnical {R}eport}. \bibinfo{institution}{SAE Technical Paper}.
\newblock
\urldef\tempurl%
\url{https://www.sae.org/publications/technical-papers/content/680010/}
\showURL{%
\tempurl}


\bibitem[Bhunia et~al\mbox{.}(2022)]%
        {bhunia2022doodleformer}
\bibfield{author}{\bibinfo{person}{Ankan~Kumar Bhunia}, \bibinfo{person}{Salman Khan}, \bibinfo{person}{Hisham Cholakkal}, \bibinfo{person}{Rao~Muhammad Anwer}, \bibinfo{person}{Fahad~Shahbaz Khan}, \bibinfo{person}{Jorma Laaksonen}, {and} \bibinfo{person}{Michael Felsberg}.} \bibinfo{year}{2022}\natexlab{}.
\newblock \showarticletitle{{DoodleFormer}: Creative Sketch Drawing with Transformers}. In \bibinfo{booktitle}{\emph{European Conference on Computer Vision}}. \bibinfo{pages}{338--355}.
\newblock
\urldef\tempurl%
\url{https://arxiv.org/abs/2112.03258}
\showURL{%
\tempurl}


\bibitem[Biederman(1987)]%
        {biederman1987geon}
\bibfield{author}{\bibinfo{person}{Irving Biederman}.} \bibinfo{year}{1987}\natexlab{}.
\newblock \showarticletitle{Recognition-by-Components: A Theory of Human Image Understanding}.
\newblock \bibinfo{journal}{\emph{Psychological Review}} \bibinfo{volume}{94}, \bibinfo{number}{2} (\bibinfo{year}{1987}), \bibinfo{pages}{115}.
\newblock
\urldef\tempurl%
\url{https://doi.org/10.1037/0033-295X.94.2.115}
\showURL{%
\tempurl}


\bibitem[Burgert et~al\mbox{.}(2024)]%
        {burgert2024diffusionillusions}
\bibfield{author}{\bibinfo{person}{Ryan Burgert}, \bibinfo{person}{Xiang Li}, \bibinfo{person}{Abe Leite}, \bibinfo{person}{Kanchana Ranasinghe}, {and} \bibinfo{person}{Michael Ryoo}.} \bibinfo{year}{2024}\natexlab{}.
\newblock \showarticletitle{Diffusion Illusions: Hiding Images in Plain Sight}. In \bibinfo{booktitle}{\emph{ACM SIGGRAPH 2024 Conference Papers}}. \bibinfo{pages}{1--11}.
\newblock
\urldef\tempurl%
\url{https://arxiv.org/abs/2312.03817}
\showURL{%
\tempurl}


\bibitem[Carlier et~al\mbox{.}(2020)]%
        {carlier2020deepsvg}
\bibfield{author}{\bibinfo{person}{Alexandre Carlier}, \bibinfo{person}{Martin Danelljan}, \bibinfo{person}{Alexandre Alahi}, {and} \bibinfo{person}{Radu Timofte}.} \bibinfo{year}{2020}\natexlab{}.
\newblock \showarticletitle{DeepSVG: A Hierarchical Generative Network for Vector Graphics Animation}.
\newblock \bibinfo{journal}{\emph{Advances in Neural Information Processing Systems}}  \bibinfo{volume}{33} (\bibinfo{year}{2020}), \bibinfo{pages}{16351--16361}.
\newblock


\bibitem[Casteljau(1959)]%
        {decasteljau1959}
\bibfield{author}{\bibinfo{person}{Paul~De Casteljau}.} \bibinfo{year}{1959}\natexlab{}.
\newblock \bibinfo{booktitle}{\emph{Outillages m{\'e}thodes calcul}}.
\newblock \bibinfo{type}{{T}echnical {R}eport}. \bibinfo{institution}{Andr{\'e} Citro{\"e}n Automobiles SA}.
\newblock


\bibitem[Cavanagh(2005)]%
        {cavanagh2005artist}
\bibfield{author}{\bibinfo{person}{Patrick Cavanagh}.} \bibinfo{year}{2005}\natexlab{}.
\newblock \showarticletitle{The Artist as Neuroscientist}.
\newblock \bibinfo{journal}{\emph{Nature}} \bibinfo{volume}{434}, \bibinfo{number}{7031} (\bibinfo{year}{2005}), \bibinfo{pages}{301--307}.
\newblock
\urldef\tempurl%
\url{https://doi.org/10.1038/434301a}
\showURL{%
\tempurl}


\bibitem[Chang et~al\mbox{.}(2025)]%
        {chang2025lookingglass}
\bibfield{author}{\bibinfo{person}{Pascal Chang}, \bibinfo{person}{Sergio Sancho}, \bibinfo{person}{Jingwei Tang}, \bibinfo{person}{Markus Gross}, {and} \bibinfo{person}{Vinicius Azevedo}.} \bibinfo{year}{2025}\natexlab{}.
\newblock \showarticletitle{LookingGlass: Generative Anamorphoses via Laplacian Pyramid Warping}. In \bibinfo{booktitle}{\emph{Proceedings of the Computer Vision and Pattern Recognition Conference}}. \bibinfo{pages}{24--33}.
\newblock


\bibitem[Chen et~al\mbox{.}(2024)]%
        {chen2024imagessound}
\bibfield{author}{\bibinfo{person}{Ziyang Chen}, \bibinfo{person}{Daniel Geng}, {and} \bibinfo{person}{Andrew Owens}.} \bibinfo{year}{2024}\natexlab{}.
\newblock \showarticletitle{Images That Sound: Composing Images and Sounds on a Single Canvas}. In \bibinfo{booktitle}{\emph{Advances in Neural Information Processing Systems}}, Vol.~\bibinfo{volume}{37}. \bibinfo{pages}{85045--85073}.
\newblock
\urldef\tempurl%
\url{https://arxiv.org/abs/2405.12221}
\showURL{%
\tempurl}


\bibitem[Das et~al\mbox{.}(2020)]%
        {das2020beziersketch}
\bibfield{author}{\bibinfo{person}{Ayan Das}, \bibinfo{person}{Yongxin Yang}, \bibinfo{person}{Timothy Hospedales}, \bibinfo{person}{Tao Xiang}, {and} \bibinfo{person}{Yi-Zhe Song}.} \bibinfo{year}{2020}\natexlab{}.
\newblock \showarticletitle{B{\'e}ziersketch: A generative model for scalable vector sketches}. In \bibinfo{booktitle}{\emph{European conference on computer vision}}. Springer, \bibinfo{pages}{632--647}.
\newblock


\bibitem[Debnath et~al\mbox{.}(2025)]%
        {debnath2025rasp}
\bibfield{author}{\bibinfo{person}{Soumyaratna Debnath}, \bibinfo{person}{Ashish Tiwari}, \bibinfo{person}{Kaustubh Sadekar}, {and} \bibinfo{person}{Shanmuganathan Raman}.} \bibinfo{year}{2025}\natexlab{}.
\newblock \showarticletitle{{RASP}: Revisiting {3D} Anamorphic Art for Shadow-Guided Packing of Irregular Objects}. In \bibinfo{booktitle}{\emph{Proceedings of the IEEE/CVF Conference on Computer Vision and Pattern Recognition}}. \bibinfo{pages}{5849--5858}.
\newblock
\urldef\tempurl%
\url{https://arxiv.org/abs/2504.02289}
\showURL{%
\tempurl}


\bibitem[Eitz et~al\mbox{.}(2012)]%
        {eitz2012humans}
\bibfield{author}{\bibinfo{person}{Mathias Eitz}, \bibinfo{person}{James Hays}, {and} \bibinfo{person}{Marc Alexa}.} \bibinfo{year}{2012}\natexlab{}.
\newblock \showarticletitle{How Do Humans Sketch Objects?}. In \bibinfo{booktitle}{\emph{ACM Transactions on Graphics}}, Vol.~\bibinfo{volume}{31}. \bibinfo{pages}{1--10}.
\newblock
\urldef\tempurl%
\url{https://dl.acm.org/doi/10.1145/2185520.2185540}
\showURL{%
\tempurl}


\bibitem[Fan et~al\mbox{.}(2023)]%
        {fan2023drawing}
\bibfield{author}{\bibinfo{person}{Judith~E Fan}, \bibinfo{person}{Wilma~A Bainbridge}, \bibinfo{person}{Rebecca Chamberlain}, {and} \bibinfo{person}{Jeffrey~D Wammes}.} \bibinfo{year}{2023}\natexlab{}.
\newblock \showarticletitle{Drawing as a versatile cognitive tool}.
\newblock \bibinfo{journal}{\emph{Nature Reviews Psychology}} \bibinfo{volume}{2}, \bibinfo{number}{9} (\bibinfo{year}{2023}), \bibinfo{pages}{556--568}.
\newblock


\bibitem[Feng et~al\mbox{.}(2024)]%
        {feng2024illusion3d}
\bibfield{author}{\bibinfo{person}{Yue Feng}, \bibinfo{person}{Vaibhav Sanjay}, \bibinfo{person}{Spencer Lutz}, \bibinfo{person}{Badour AlBahar}, \bibinfo{person}{Songwei Ge}, {and} \bibinfo{person}{Jia-Bin Huang}.} \bibinfo{year}{2024}\natexlab{}.
\newblock \showarticletitle{{Illusion3D}: {3D} Multiview Illusion with {2D} Diffusion Priors}.
\newblock \bibinfo{journal}{\emph{arXiv preprint arXiv:2412.09625}} (\bibinfo{year}{2024}).
\newblock
\urldef\tempurl%
\url{https://arxiv.org/abs/2412.09625}
\showURL{%
\tempurl}


\bibitem[Frans et~al\mbox{.}(2022)]%
        {frans2022clipdraw}
\bibfield{author}{\bibinfo{person}{Kevin Frans}, \bibinfo{person}{Lisa Soros}, {and} \bibinfo{person}{Olaf Witkowski}.} \bibinfo{year}{2022}\natexlab{}.
\newblock \showarticletitle{Clipdraw: Exploring text-to-drawing synthesis through language-image encoders}.
\newblock \bibinfo{journal}{\emph{Advances in Neural Information Processing Systems}}  \bibinfo{volume}{35} (\bibinfo{year}{2022}), \bibinfo{pages}{5207--5218}.
\newblock
\urldef\tempurl%
\url{https://arxiv.org/abs/2106.14843}
\showURL{%
\tempurl}


\bibitem[Gao et~al\mbox{.}(2025)]%
        {gao2025ptdiffusion}
\bibfield{author}{\bibinfo{person}{Xiang Gao}, \bibinfo{person}{Shuai Yang}, {and} \bibinfo{person}{Jiaying Liu}.} \bibinfo{year}{2025}\natexlab{}.
\newblock \showarticletitle{PTDiffusion: Free Lunch for Generating Optical Illusion Hidden Pictures with Phase-Transferred Diffusion Model}. In \bibinfo{booktitle}{\emph{Proceedings of the IEEE/CVF Conference on Computer Vision and Pattern Recognition}}. \bibinfo{pages}{18240--18249}.
\newblock


\bibitem[Geng et~al\mbox{.}(2024a)]%
        {geng2024factorized}
\bibfield{author}{\bibinfo{person}{Daniel Geng}, \bibinfo{person}{Inbum Park}, {and} \bibinfo{person}{Andrew Owens}.} \bibinfo{year}{2024}\natexlab{a}.
\newblock \showarticletitle{Factorized Diffusion: Perceptual Illusions by Noise Decomposition}. In \bibinfo{booktitle}{\emph{European Conference on Computer Vision}}. \bibinfo{pages}{366--384}.
\newblock
\urldef\tempurl%
\url{https://arxiv.org/abs/2404.11615}
\showURL{%
\tempurl}


\bibitem[Geng et~al\mbox{.}(2024b)]%
        {geng2024visualanagrams}
\bibfield{author}{\bibinfo{person}{Daniel Geng}, \bibinfo{person}{Inbum Park}, {and} \bibinfo{person}{Andrew Owens}.} \bibinfo{year}{2024}\natexlab{b}.
\newblock \showarticletitle{Visual Anagrams: Generating Multi-View Optical Illusions with Diffusion Models}. In \bibinfo{booktitle}{\emph{Proceedings of the IEEE/CVF Conference on Computer Vision and Pattern Recognition}}. \bibinfo{pages}{24154--24163}.
\newblock
\urldef\tempurl%
\url{https://arxiv.org/abs/2311.17919}
\showURL{%
\tempurl}


\bibitem[Gregor et~al\mbox{.}(2015)]%
        {gregor2015draw}
\bibfield{author}{\bibinfo{person}{Karol Gregor}, \bibinfo{person}{Ivo Danihelka}, \bibinfo{person}{Alex Graves}, \bibinfo{person}{Danilo Rezende}, {and} \bibinfo{person}{Daan Wierstra}.} \bibinfo{year}{2015}\natexlab{}.
\newblock \showarticletitle{Draw: A recurrent neural network for image generation}. In \bibinfo{booktitle}{\emph{International conference on machine learning}}. PMLR, \bibinfo{pages}{1462--1471}.
\newblock


\bibitem[Ha and Eck(2017)]%
        {ha2017neural}
\bibfield{author}{\bibinfo{person}{David Ha} {and} \bibinfo{person}{Douglas Eck}.} \bibinfo{year}{2017}\natexlab{}.
\newblock \showarticletitle{A Neural Representation of Sketch Drawings}.
\newblock \bibinfo{journal}{\emph{arXiv preprint arXiv:1704.03477}} (\bibinfo{year}{2017}).
\newblock
\urldef\tempurl%
\url{https://arxiv.org/abs/1704.03477}
\showURL{%
\tempurl}


\bibitem[Hessel et~al\mbox{.}(2022)]%
        {hessel2022clipscorereferencefreeevaluationmetric}
\bibfield{author}{\bibinfo{person}{Jack Hessel}, \bibinfo{person}{Ari Holtzman}, \bibinfo{person}{Maxwell Forbes}, \bibinfo{person}{Ronan~Le Bras}, {and} \bibinfo{person}{Yejin Choi}.} \bibinfo{year}{2022}\natexlab{}.
\newblock \bibinfo{title}{CLIPScore: A Reference-free Evaluation Metric for Image Captioning}.
\newblock
\showeprint[arxiv]{2104.08718}~[cs.CV]
\urldef\tempurl%
\url{https://arxiv.org/abs/2104.08718}
\showURL{%
\tempurl}


\bibitem[Hsiao et~al\mbox{.}(2018)]%
        {hsiao2018wireart}
\bibfield{author}{\bibinfo{person}{Kai-Wen Hsiao}, \bibinfo{person}{Jia-Bin Huang}, {and} \bibinfo{person}{Hung-Kuo Chu}.} \bibinfo{year}{2018}\natexlab{}.
\newblock \showarticletitle{Multi-View Wire Art}. In \bibinfo{booktitle}{\emph{ACM Transactions on Graphics}}, Vol.~\bibinfo{volume}{37}. \bibinfo{pages}{242}.
\newblock
\urldef\tempurl%
\url{https://dl.acm.org/doi/10.1145/3272127.3275087}
\showURL{%
\tempurl}


\bibitem[Jain et~al\mbox{.}(2023)]%
        {jain2023vectorfusion}
\bibfield{author}{\bibinfo{person}{Ajay Jain}, \bibinfo{person}{Amber Xie}, {and} \bibinfo{person}{Pieter Abbeel}.} \bibinfo{year}{2023}\natexlab{}.
\newblock \showarticletitle{VectorFusion: Text-to-SVG by Abstracting Pixel-Based Diffusion Models}. In \bibinfo{booktitle}{\emph{Proceedings of the IEEE/CVF Conference on Computer Vision and Pattern Recognition}}. \bibinfo{pages}{1911--1920}.
\newblock


\bibitem[Jongejan et~al\mbox{.}(2016)]%
        {jongejan2016quickdraw}
\bibfield{author}{\bibinfo{person}{Jonas Jongejan}, \bibinfo{person}{Henry Rowley}, \bibinfo{person}{Takashi Kawashima}, \bibinfo{person}{Jongmin Kim}, {and} \bibinfo{person}{Nick Fox-Gieg}.} \bibinfo{year}{2016}\natexlab{}.
\newblock \bibinfo{title}{Quick, {Draw}! The Data}.
\newblock
\urldef\tempurl%
\url{https://quickdraw.withgoogle.com/data}
\showURL{%
\tempurl}


\bibitem[Kanizsa et~al\mbox{.}(1979)]%
        {kanizsa1979}
\bibfield{author}{\bibinfo{person}{Gaetano Kanizsa}, \bibinfo{person}{Paolo Legrenzi}, {and} \bibinfo{person}{Paolo Bozzi}.} \bibinfo{year}{1979}\natexlab{}.
\newblock \bibinfo{booktitle}{\emph{Organization in Vision: Essays on {Gestalt} Perception}}.
\newblock \bibinfo{publisher}{Praeger}.
\newblock


\bibitem[Li et~al\mbox{.}(2020)]%
        {li2020differentiable}
\bibfield{author}{\bibinfo{person}{Tzu-Mao Li}, \bibinfo{person}{Michal Luk{\'a}{\v{c}}}, \bibinfo{person}{Micha{\"e}l Gharbi}, {and} \bibinfo{person}{Jonathan Ragan-Kelley}.} \bibinfo{year}{2020}\natexlab{}.
\newblock \showarticletitle{Differentiable vector graphics rasterization for editing and learning}.
\newblock \bibinfo{journal}{\emph{ACM Transactions on Graphics (TOG)}} \bibinfo{volume}{39}, \bibinfo{number}{6} (\bibinfo{year}{2020}), \bibinfo{pages}{1--15}.
\newblock
\urldef\tempurl%
\url{https://people.csail.mit.edu/tzumao/diffvg/}
\showURL{%
\tempurl}


\bibitem[Liang et~al\mbox{.}(2024)]%
        {liang2024luciddreamer}
\bibfield{author}{\bibinfo{person}{Yixun Liang}, \bibinfo{person}{Xin Yang}, \bibinfo{person}{Jiantao Lin}, \bibinfo{person}{Haodong Li}, \bibinfo{person}{Xiaogang Xu}, {and} \bibinfo{person}{Yingcong Chen}.} \bibinfo{year}{2024}\natexlab{}.
\newblock \showarticletitle{{LucidDreamer}: Towards High-Fidelity Text-to-{3D} Generation via Interval Score Matching}. In \bibinfo{booktitle}{\emph{Proceedings of the IEEE/CVF Conference on Computer Vision and Pattern Recognition}}. \bibinfo{pages}{6517--6526}.
\newblock
\urldef\tempurl%
\url{https://arxiv.org/abs/2311.11284}
\showURL{%
\tempurl}


\bibitem[Liu et~al\mbox{.}(2019)]%
        {liu2019sketchgan}
\bibfield{author}{\bibinfo{person}{Fang Liu}, \bibinfo{person}{Xiaoming Deng}, \bibinfo{person}{Yu-Kun Lai}, \bibinfo{person}{Yong-Jin Liu}, \bibinfo{person}{Cuixia Ma}, {and} \bibinfo{person}{Hongan Wang}.} \bibinfo{year}{2019}\natexlab{}.
\newblock \showarticletitle{{SketchGAN}: Joint Sketch Completion and Recognition with Generative Adversarial Network}. In \bibinfo{booktitle}{\emph{Proceedings of the IEEE/CVF Conference on Computer Vision and Pattern Recognition}}. \bibinfo{pages}{5830--5839}.
\newblock


\bibitem[Lukoianov et~al\mbox{.}(2024)]%
        {lukoianov2024ddim}
\bibfield{author}{\bibinfo{person}{Artem Lukoianov}, \bibinfo{person}{Haitz~S{\'a}ez de Oc{\'a}riz~Borde}, \bibinfo{person}{Kristjan Greenewald}, \bibinfo{person}{Vitor Guizilini}, \bibinfo{person}{Timur Bagautdinov}, \bibinfo{person}{Vincent Sitzmann}, {and} \bibinfo{person}{Justin~M. Solomon}.} \bibinfo{year}{2024}\natexlab{}.
\newblock \showarticletitle{Score Distillation via Reparametrized {DDIM}}. In \bibinfo{booktitle}{\emph{Advances in Neural Information Processing Systems}}, Vol.~\bibinfo{volume}{37}. \bibinfo{pages}{26011--26044}.
\newblock
\urldef\tempurl%
\url{https://arxiv.org/abs/2405.15891}
\showURL{%
\tempurl}


\bibitem[Luo et~al\mbox{.}(2025)]%
        {luo2025shadowdraw}
\bibfield{author}{\bibinfo{person}{Rundong Luo}, \bibinfo{person}{Noah Snavely}, {and} \bibinfo{person}{Wei-Chiu Ma}.} \bibinfo{year}{2025}\natexlab{}.
\newblock \showarticletitle{{ShadowDraw}: From Any Object to Shadow-Drawing Compositional Art}.
\newblock \bibinfo{journal}{\emph{arXiv preprint arXiv:2512.05110}} (\bibinfo{year}{2025}).
\newblock
\urldef\tempurl%
\url{https://arxiv.org/abs/2512.05110}
\showURL{%
\tempurl}


\bibitem[Mitra and Pauly(2009)]%
        {mitra2009shadowart}
\bibfield{author}{\bibinfo{person}{Niloy~J. Mitra} {and} \bibinfo{person}{Mark Pauly}.} \bibinfo{year}{2009}\natexlab{}.
\newblock \showarticletitle{Shadow Art}. In \bibinfo{booktitle}{\emph{ACM Transactions on Graphics}}, Vol.~\bibinfo{volume}{28}. \bibinfo{pages}{156}.
\newblock
\urldef\tempurl%
\url{https://dl.acm.org/doi/10.1145/1618452.1618502}
\showURL{%
\tempurl}


\bibitem[Oliva et~al\mbox{.}(2006)]%
        {oliva2006hybrid}
\bibfield{author}{\bibinfo{person}{Aude Oliva}, \bibinfo{person}{Antonio Torralba}, {and} \bibinfo{person}{Philippe~G. Schyns}.} \bibinfo{year}{2006}\natexlab{}.
\newblock \showarticletitle{Hybrid Images}. In \bibinfo{booktitle}{\emph{ACM Transactions on Graphics}}, Vol.~\bibinfo{volume}{25}. \bibinfo{pages}{527--532}.
\newblock
\urldef\tempurl%
\url{https://dl.acm.org/doi/10.1145/1141911.1141951}
\showURL{%
\tempurl}


\bibitem[Podell et~al\mbox{.}(2023)]%
        {podell2023sdxlimprovinglatentdiffusion}
\bibfield{author}{\bibinfo{person}{Dustin Podell}, \bibinfo{person}{Zion English}, \bibinfo{person}{Kyle Lacey}, \bibinfo{person}{Andreas Blattmann}, \bibinfo{person}{Tim Dockhorn}, \bibinfo{person}{Jonas Müller}, \bibinfo{person}{Joe Penna}, {and} \bibinfo{person}{Robin Rombach}.} \bibinfo{year}{2023}\natexlab{}.
\newblock \bibinfo{title}{SDXL: Improving Latent Diffusion Models for High-Resolution Image Synthesis}.
\newblock
\showeprint[arxiv]{2307.01952}~[cs.CV]
\urldef\tempurl%
\url{https://arxiv.org/abs/2307.01952}
\showURL{%
\tempurl}


\bibitem[Polaczek et~al\mbox{.}(2025)]%
        {polaczek2025neuralsvg}
\bibfield{author}{\bibinfo{person}{Sagi Polaczek}, \bibinfo{person}{Yuval Alaluf}, \bibinfo{person}{Elad Richardson}, \bibinfo{person}{Yael Vinker}, {and} \bibinfo{person}{Daniel Cohen-Or}.} \bibinfo{year}{2025}\natexlab{}.
\newblock \showarticletitle{Neuralsvg: An implicit representation for text-to-vector generation}.
\newblock \bibinfo{journal}{\emph{arXiv preprint arXiv:2501.03992}} (\bibinfo{year}{2025}).
\newblock


\bibitem[Poole et~al\mbox{.}(2022)]%
        {poole2022dreamfusion}
\bibfield{author}{\bibinfo{person}{Ben Poole}, \bibinfo{person}{Ajay Jain}, \bibinfo{person}{Jonathan~T. Barron}, {and} \bibinfo{person}{Ben Mildenhall}.} \bibinfo{year}{2022}\natexlab{}.
\newblock \showarticletitle{{DreamFusion}: Text-to-{3D} Using {2D} Diffusion}.
\newblock \bibinfo{journal}{\emph{arXiv preprint arXiv:2209.14988}} (\bibinfo{year}{2022}).
\newblock
\urldef\tempurl%
\url{https://arxiv.org/abs/2209.14988}
\showURL{%
\tempurl}


\bibitem[Pratt et~al\mbox{.}(2023)]%
        {pratt2023anamorphic}
\bibfield{author}{\bibinfo{person}{Louis Pratt}, \bibinfo{person}{Andrew Johnston}, {and} \bibinfo{person}{Nico Pietroni}.} \bibinfo{year}{2023}\natexlab{}.
\newblock \showarticletitle{Bending the Light: Next Generation Anamorphic Sculptures}.
\newblock \bibinfo{journal}{\emph{Computers \& Graphics}}  \bibinfo{volume}{114} (\bibinfo{year}{2023}), \bibinfo{pages}{210--218}.
\newblock


\bibitem[Qu et~al\mbox{.}(2023)]%
        {qu2023sketchdreamer}
\bibfield{author}{\bibinfo{person}{Zhiyu Qu}, \bibinfo{person}{Tao Xiang}, {and} \bibinfo{person}{Yi-Zhe Song}.} \bibinfo{year}{2023}\natexlab{}.
\newblock \showarticletitle{{SketchDreamer}: Interactive Text-Augmented Creative Sketch Ideation}.
\newblock \bibinfo{journal}{\emph{arXiv preprint arXiv:2308.14191}}.
\newblock
\urldef\tempurl%
\url{https://arxiv.org/abs/2308.14191}
\showURL{%
\tempurl}


\bibitem[Radford et~al\mbox{.}(2021)]%
        {radford2021learning}
\bibfield{author}{\bibinfo{person}{Alec Radford}, \bibinfo{person}{Jong~Wook Kim}, \bibinfo{person}{Chris Hallacy}, \bibinfo{person}{Aditya Ramesh}, \bibinfo{person}{Gabriel Goh}, \bibinfo{person}{Sandhini Agarwal}, \bibinfo{person}{Girish Sastry}, \bibinfo{person}{Amanda Askell}, \bibinfo{person}{Pamela Mishkin}, \bibinfo{person}{Jack Clark}, {et~al\mbox{.}}} \bibinfo{year}{2021}\natexlab{}.
\newblock \showarticletitle{Learning Transferable Visual Models from Natural Language Supervision}. In \bibinfo{booktitle}{\emph{International Conference on Machine Learning (ICML)}}. \bibinfo{pages}{8748--8763}.
\newblock
\urldef\tempurl%
\url{https://arxiv.org/abs/2103.00020}
\showURL{%
\tempurl}


\bibitem[Reddy et~al\mbox{.}(2021)]%
        {reddy2021im2vec}
\bibfield{author}{\bibinfo{person}{Pradyumna Reddy}, \bibinfo{person}{Michael Gharbi}, \bibinfo{person}{Michal Lukac}, {and} \bibinfo{person}{Niloy~J. Mitra}.} \bibinfo{year}{2021}\natexlab{}.
\newblock \showarticletitle{{Im2Vec}: Synthesizing Vector Graphics Without Vector Supervision}. In \bibinfo{booktitle}{\emph{Proceedings of the IEEE/CVF Conference on Computer Vision and Pattern Recognition}}. \bibinfo{pages}{7342--7351}.
\newblock
\urldef\tempurl%
\url{https://arxiv.org/abs/2102.02798}
\showURL{%
\tempurl}


\bibitem[Ribeiro et~al\mbox{.}(2020)]%
        {ribeiro2020sketchformer}
\bibfield{author}{\bibinfo{person}{Leo Sampaio~Ferraz Ribeiro}, \bibinfo{person}{Tu Bui}, \bibinfo{person}{John Collomosse}, {and} \bibinfo{person}{Moacir Ponti}.} \bibinfo{year}{2020}\natexlab{}.
\newblock \showarticletitle{Sketchformer: Transformer-Based Representation for Sketched Structure}. In \bibinfo{booktitle}{\emph{Proceedings of the IEEE/CVF conference on computer vision and pattern recognition}}. \bibinfo{pages}{14153--14162}.
\newblock


\bibitem[Rodriguez et~al\mbox{.}(2025)]%
        {rodriguez2025starvector}
\bibfield{author}{\bibinfo{person}{Juan~A Rodriguez}, \bibinfo{person}{Abhay Puri}, \bibinfo{person}{Shubham Agarwal}, \bibinfo{person}{Issam~H Laradji}, \bibinfo{person}{Pau Rodriguez}, \bibinfo{person}{Sai Rajeswar}, \bibinfo{person}{David Vazquez}, \bibinfo{person}{Christopher Pal}, {and} \bibinfo{person}{Marco Pedersoli}.} \bibinfo{year}{2025}\natexlab{}.
\newblock \showarticletitle{Starvector: Generating scalable vector graphics code from images and text}. In \bibinfo{booktitle}{\emph{Proceedings of the Computer Vision and Pattern Recognition Conference}}. \bibinfo{pages}{16175--16186}.
\newblock


\bibitem[Rombach et~al\mbox{.}(2022)]%
        {rombach2022ldm}
\bibfield{author}{\bibinfo{person}{Robin Rombach}, \bibinfo{person}{Andreas Blattmann}, \bibinfo{person}{Dominik Lorenz}, \bibinfo{person}{Patrick Esser}, {and} \bibinfo{person}{Bj{\"o}rn Ommer}.} \bibinfo{year}{2022}\natexlab{}.
\newblock \showarticletitle{High-resolution image synthesis with latent diffusion models}. In \bibinfo{booktitle}{\emph{Proceedings of the IEEE/CVF conference on computer vision and pattern recognition}}. \bibinfo{pages}{10684--10695}.
\newblock
\urldef\tempurl%
\url{https://arxiv.org/abs/2112.10752}
\showURL{%
\tempurl}


\bibitem[Sangkloy et~al\mbox{.}(2016)]%
        {sangkloy2016sketchy}
\bibfield{author}{\bibinfo{person}{Patsorn Sangkloy}, \bibinfo{person}{Nathan Burnell}, \bibinfo{person}{Cusuh Ham}, {and} \bibinfo{person}{James Hays}.} \bibinfo{year}{2016}\natexlab{}.
\newblock \showarticletitle{The sketchy database: learning to retrieve badly drawn bunnies}.
\newblock \bibinfo{journal}{\emph{ACM Trans. Graph.}} \bibinfo{volume}{35}, \bibinfo{number}{4} (\bibinfo{year}{2016}).
\newblock
\urldef\tempurl%
\url{https://doi.org/10.1145/2897824.2925954}
\showURL{%
\tempurl}


\bibitem[Su et~al\mbox{.}(2020)]%
        {su2020sketchhealer}
\bibfield{author}{\bibinfo{person}{Guoyao Su}, \bibinfo{person}{Yonggang Qi}, \bibinfo{person}{Kaiyue Pang}, \bibinfo{person}{Jie Yang}, {and} \bibinfo{person}{Yi-Zhe Song}.} \bibinfo{year}{2020}\natexlab{}.
\newblock \showarticletitle{{SketchHealer}: A Graph-to-Sequence Network for Recreating Partial Human Sketches}. In \bibinfo{booktitle}{\emph{Proceedings of The 31st British Machine Vision Conference (BMVC)}}.
\newblock


\bibitem[Thamizharasan et~al\mbox{.}(2024)]%
        {thamizharasan2024nivel}
\bibfield{author}{\bibinfo{person}{Vikas Thamizharasan}, \bibinfo{person}{Difan Liu}, \bibinfo{person}{Matthew Fisher}, \bibinfo{person}{Nanxuan Zhao}, \bibinfo{person}{Evangelos Kalogerakis}, {and} \bibinfo{person}{Michal Lukac}.} \bibinfo{year}{2024}\natexlab{}.
\newblock \showarticletitle{Nivel: Neural implicit vector layers for text-to-vector generation}. In \bibinfo{booktitle}{\emph{Proceedings of the IEEE/CVF Conference on Computer Vision and Pattern Recognition}}. \bibinfo{pages}{4589--4597}.
\newblock


\bibitem[Vinker et~al\mbox{.}(2023)]%
        {vinker2023clipascene}
\bibfield{author}{\bibinfo{person}{Yael Vinker}, \bibinfo{person}{Yuval Alaluf}, \bibinfo{person}{Daniel Cohen-Or}, {and} \bibinfo{person}{Ariel Shamir}.} \bibinfo{year}{2023}\natexlab{}.
\newblock \showarticletitle{CLIPascene: Scene Sketching with Different Types and Levels of Abstraction}. In \bibinfo{booktitle}{\emph{Proceedings of the IEEE/CVF International Conference on Computer Vision}}. \bibinfo{pages}{4146--4156}.
\newblock
\urldef\tempurl%
\url{https://arxiv.org/abs/2211.17256}
\showURL{%
\tempurl}


\bibitem[Vinker et~al\mbox{.}(2022)]%
        {vinker2022clipasso}
\bibfield{author}{\bibinfo{person}{Yael Vinker}, \bibinfo{person}{Ehsan Pajouheshgar}, \bibinfo{person}{Jessica~Y Bo}, \bibinfo{person}{Roman~Christian Bachmann}, \bibinfo{person}{Amit~Haim Bermano}, \bibinfo{person}{Daniel Cohen-Or}, \bibinfo{person}{Amir Zamir}, {and} \bibinfo{person}{Ariel Shamir}.} \bibinfo{year}{2022}\natexlab{}.
\newblock \showarticletitle{Clipasso: Semantically-aware object sketching}.
\newblock \bibinfo{journal}{\emph{ACM Transactions on Graphics}} \bibinfo{volume}{41}, \bibinfo{number}{4} (\bibinfo{year}{2022}), \bibinfo{pages}{1--11}.
\newblock


\bibitem[Vinker et~al\mbox{.}(2025)]%
        {vinker2025sketchagent}
\bibfield{author}{\bibinfo{person}{Yael Vinker}, \bibinfo{person}{Tamar~Rott Shaham}, \bibinfo{person}{Kristine Zheng}, \bibinfo{person}{Alex Zhao}, \bibinfo{person}{Judith E~Fan}, {and} \bibinfo{person}{Antonio Torralba}.} \bibinfo{year}{2025}\natexlab{}.
\newblock \showarticletitle{{SketchAgent}: Language-Driven Sequential Sketch Generation}. In \bibinfo{booktitle}{\emph{Proceedings of the Computer Vision and Pattern Recognition Conference}}. \bibinfo{pages}{23355--23368}.
\newblock


\bibitem[Wagemans et~al\mbox{.}(2012)]%
        {wagemans2012gestalt}
\bibfield{author}{\bibinfo{person}{Johan Wagemans}, \bibinfo{person}{James~H. Elder}, \bibinfo{person}{Michael Kubovy}, \bibinfo{person}{Stephen~E. Palmer}, \bibinfo{person}{Mary~A. Peterson}, \bibinfo{person}{Manish Singh}, {and} \bibinfo{person}{R{\"u}diger~Von der Heydt}.} \bibinfo{year}{2012}\natexlab{}.
\newblock \showarticletitle{A Century of {Gestalt} Psychology in Visual Perception: {I}. Perceptual Grouping and Figure--Ground Organization}.
\newblock \bibinfo{journal}{\emph{Psychological Bulletin}} \bibinfo{volume}{138}, \bibinfo{number}{6} (\bibinfo{year}{2012}), \bibinfo{pages}{1172}.
\newblock


\bibitem[Wang et~al\mbox{.}(2023a)]%
        {wang2023sketchknitter}
\bibfield{author}{\bibinfo{person}{Qiang Wang}, \bibinfo{person}{Haoge Deng}, \bibinfo{person}{Yonggang Qi}, \bibinfo{person}{Da Li}, {and} \bibinfo{person}{Yi-Zhe Song}.} \bibinfo{year}{2023}\natexlab{a}.
\newblock \showarticletitle{{SketchKnitter}: Vectorized Sketch Generation with Diffusion Models}. In \bibinfo{booktitle}{\emph{International Conference on Learning Representations}}.
\newblock
\urldef\tempurl%
\url{https://openreview.net/forum?id=4eJ43EN2g6l}
\showURL{%
\tempurl}


\bibitem[Wang et~al\mbox{.}(2023b)]%
        {wang2023prolificdreamer}
\bibfield{author}{\bibinfo{person}{Zhengyi Wang}, \bibinfo{person}{Cheng Lu}, \bibinfo{person}{Yikai Wang}, \bibinfo{person}{Fan Bao}, \bibinfo{person}{Chongxuan Li}, \bibinfo{person}{Hang Su}, {and} \bibinfo{person}{Jun Zhu}.} \bibinfo{year}{2023}\natexlab{b}.
\newblock \showarticletitle{{ProlificDreamer}: High-Fidelity and Diverse Text-to-{3D} Generation with Variational Score Distillation}. In \bibinfo{booktitle}{\emph{Advances in Neural Information Processing Systems}}, Vol.~\bibinfo{volume}{36}. \bibinfo{pages}{8406--8441}.
\newblock
\urldef\tempurl%
\url{https://arxiv.org/abs/2305.16213}
\showURL{%
\tempurl}


\bibitem[Wu et~al\mbox{.}(2022)]%
        {wu2022mirror}
\bibfield{author}{\bibinfo{person}{Kang Wu}, \bibinfo{person}{Renjie Chen}, \bibinfo{person}{Xiao-Ming Fu}, {and} \bibinfo{person}{Ligang Liu}.} \bibinfo{year}{2022}\natexlab{}.
\newblock \showarticletitle{Computational Mirror Cup and Saucer Art}. In \bibinfo{booktitle}{\emph{ACM Transactions on Graphics}}, Vol.~\bibinfo{volume}{41}. \bibinfo{pages}{1--15}.
\newblock
\urldef\tempurl%
\url{https://dl.acm.org/doi/10.1145/3516428}
\showURL{%
\tempurl}


\bibitem[Wu et~al\mbox{.}(2025)]%
        {wu2025chat2svg}
\bibfield{author}{\bibinfo{person}{Ronghuan Wu}, \bibinfo{person}{Wanchao Su}, {and} \bibinfo{person}{Jing Liao}.} \bibinfo{year}{2025}\natexlab{}.
\newblock \showarticletitle{Chat2SVG: Vector Graphics Generation with Large Language Models and Image Diffusion Models}. In \bibinfo{booktitle}{\emph{Proceedings of the Computer Vision and Pattern Recognition Conference}}. \bibinfo{pages}{23690--23700}.
\newblock


\bibitem[Wu et~al\mbox{.}(2023b)]%
        {wu2023iconshop}
\bibfield{author}{\bibinfo{person}{Ronghuan Wu}, \bibinfo{person}{Wanchao Su}, \bibinfo{person}{Kede Ma}, {and} \bibinfo{person}{Jing Liao}.} \bibinfo{year}{2023}\natexlab{b}.
\newblock \showarticletitle{Iconshop: Text-guided vector icon synthesis with autoregressive transformers}.
\newblock \bibinfo{journal}{\emph{ACM Transactions on Graphics (TOG)}} \bibinfo{volume}{42}, \bibinfo{number}{6} (\bibinfo{year}{2023}), \bibinfo{pages}{1--14}.
\newblock


\bibitem[Wu et~al\mbox{.}(2023a)]%
        {wu2023human}
\bibfield{author}{\bibinfo{person}{Xiaoshi Wu}, \bibinfo{person}{Yiming Hao}, \bibinfo{person}{Keqiang Sun}, \bibinfo{person}{Yixiong Chen}, \bibinfo{person}{Feng Zhu}, \bibinfo{person}{Rui Zhao}, {and} \bibinfo{person}{Hongsheng Li}.} \bibinfo{year}{2023}\natexlab{a}.
\newblock \showarticletitle{Human Preference Score v2: A Solid Benchmark for Evaluating Human Preferences of Text-to-Image Synthesis}.
\newblock \bibinfo{journal}{\emph{arXiv preprint arXiv:2306.09341}} (\bibinfo{year}{2023}).
\newblock


\bibitem[Xing et~al\mbox{.}(2025)]%
        {xing2025llmsvg}
\bibfield{author}{\bibinfo{person}{Ximing Xing}, \bibinfo{person}{Juncheng Hu}, \bibinfo{person}{Jing Zhang}, \bibinfo{person}{Dong Xu}, {and} \bibinfo{person}{Qian Yu}.} \bibinfo{year}{2025}\natexlab{}.
\newblock \showarticletitle{Empowering {LLMs} to Understand and Generate Complex Vector Graphics}. In \bibinfo{booktitle}{\emph{Proceedings of the IEEE/CVF Conference on Computer Vision and Pattern Recognition}}. \bibinfo{pages}{19487--19497}.
\newblock
\urldef\tempurl%
\url{https://arxiv.org/abs/2412.11102}
\showURL{%
\tempurl}


\bibitem[Xing et~al\mbox{.}(2023)]%
        {xing2023diffsketcher}
\bibfield{author}{\bibinfo{person}{XiMing Xing}, \bibinfo{person}{Chuang Wang}, \bibinfo{person}{Haitao Zhou}, \bibinfo{person}{Jing Zhang}, \bibinfo{person}{Qian Yu}, {and} \bibinfo{person}{Dong Xu}.} \bibinfo{year}{2023}\natexlab{}.
\newblock \showarticletitle{DiffSketcher: Text Guided Vector Sketch Synthesis through Latent Diffusion Models}. In \bibinfo{booktitle}{\emph{Thirty-seventh Conference on Neural Information Processing Systems}}.
\newblock
\urldef\tempurl%
\url{https://openreview.net/forum?id=CY1xatvEQj}
\showURL{%
\tempurl}


\bibitem[Xing et~al\mbox{.}(2024)]%
        {xing2024svgdreamer}
\bibfield{author}{\bibinfo{person}{Ximing Xing}, \bibinfo{person}{Haitao Zhou}, \bibinfo{person}{Chuang Wang}, \bibinfo{person}{Jing Zhang}, \bibinfo{person}{Dong Xu}, {and} \bibinfo{person}{Qian Yu}.} \bibinfo{year}{2024}\natexlab{}.
\newblock \showarticletitle{Svgdreamer: Text guided svg generation with diffusion model}. In \bibinfo{booktitle}{\emph{Proceedings of the IEEE/CVF Conference on Computer Vision and Pattern Recognition}}. \bibinfo{pages}{4546--4555}.
\newblock


\bibitem[Xu et~al\mbox{.}(2023)]%
        {xu2023imagereward}
\bibfield{author}{\bibinfo{person}{Jiazheng Xu}, \bibinfo{person}{Xiao Liu}, \bibinfo{person}{Yuchen Wu}, \bibinfo{person}{Yuxuan Tong}, \bibinfo{person}{Qinkai Li}, \bibinfo{person}{Ming Ding}, \bibinfo{person}{Jie Tang}, {and} \bibinfo{person}{Yuxiao Dong}.} \bibinfo{year}{2023}\natexlab{}.
\newblock \showarticletitle{ImageReward: learning and evaluating human preferences for text-to-image generation}. In \bibinfo{booktitle}{\emph{Proceedings of the 37th International Conference on Neural Information Processing Systems}}. \bibinfo{pages}{15903--15935}.
\newblock


\bibitem[Yu et~al\mbox{.}(2017)]%
        {yu2017sketchanet}
\bibfield{author}{\bibinfo{person}{Qian Yu}, \bibinfo{person}{Yongxin Yang}, \bibinfo{person}{Feng Liu}, \bibinfo{person}{Yi-Zhe Song}, \bibinfo{person}{Tao Xiang}, {and} \bibinfo{person}{Timothy~M. Hospedales}.} \bibinfo{year}{2017}\natexlab{}.
\newblock \showarticletitle{Sketch-a-Net: A Deep Neural Network That Beats Humans}.
\newblock \bibinfo{journal}{\emph{International Journal of Computer Vision}} \bibinfo{volume}{122}, \bibinfo{number}{3}, \bibinfo{pages}{411--425}.
\newblock
\urldef\tempurl%
\url{https://arxiv.org/abs/1501.07873}
\showURL{%
\tempurl}


\bibitem[Zang et~al\mbox{.}(2025)]%
        {zang2025hierarchical}
\bibfield{author}{\bibinfo{person}{Sicong Zang}, \bibinfo{person}{Shuhui Gao}, {and} \bibinfo{person}{Zhijun Fang}.} \bibinfo{year}{2025}\natexlab{}.
\newblock \showarticletitle{Generating Sketches in a Hierarchical Auto-Regressive Process for Flexible Sketch Drawing Manipulation at Stroke-Level}.
\newblock \bibinfo{journal}{\emph{arXiv preprint arXiv:2511.07889}} (\bibinfo{year}{2025}).
\newblock


\bibitem[Zhao et~al\mbox{.}(2023)]%
        {zhao2023ambigen}
\bibfield{author}{\bibinfo{person}{Boheng Zhao}, \bibinfo{person}{Rana Hanocka}, {and} \bibinfo{person}{Raymond~A. Yeh}.} \bibinfo{year}{2023}\natexlab{}.
\newblock \showarticletitle{{AmbiGen}: Generating Ambigrams from Pre-Trained Diffusion Model}.
\newblock \bibinfo{journal}{\emph{arXiv preprint arXiv:2312.02967}} (\bibinfo{year}{2023}).
\newblock
\urldef\tempurl%
\url{https://arxiv.org/abs/2312.02967}
\showURL{%
\tempurl}


\end{thebibliography}

%%
%% If your work has an appendix, this is the place to put it.
\appendix

\begin{figure*}[t]
  \centering
  \includegraphics[width=0.97\textwidth]{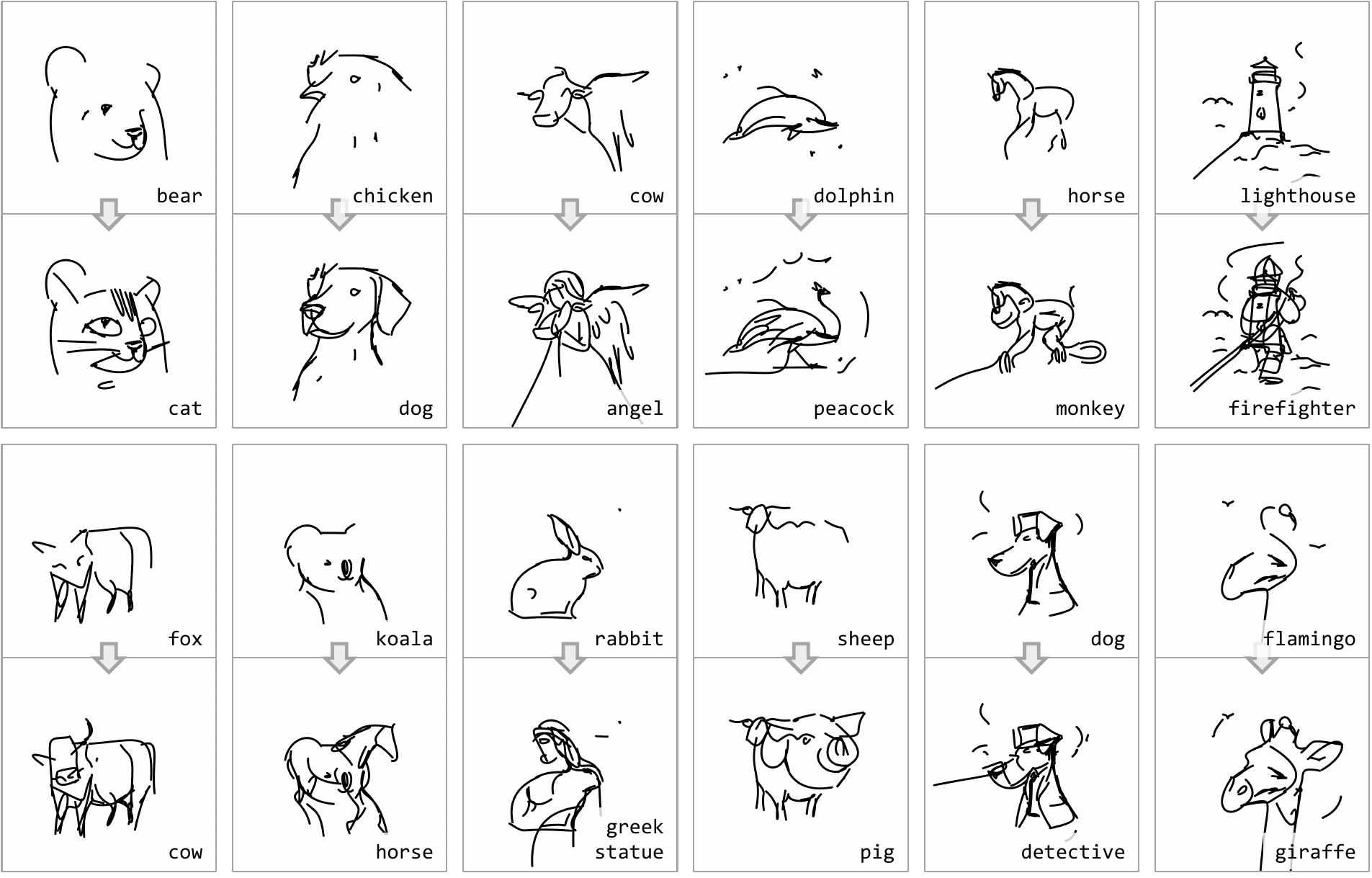}
  \caption{\textbf{Additional 2-phase progressive illusion results produced by our method.}
  }
  \label{fig:appendix_2_phase}
\end{figure*}

\begin{figure*}[t]
  \centering
  \includegraphics[width=0.97\textwidth]{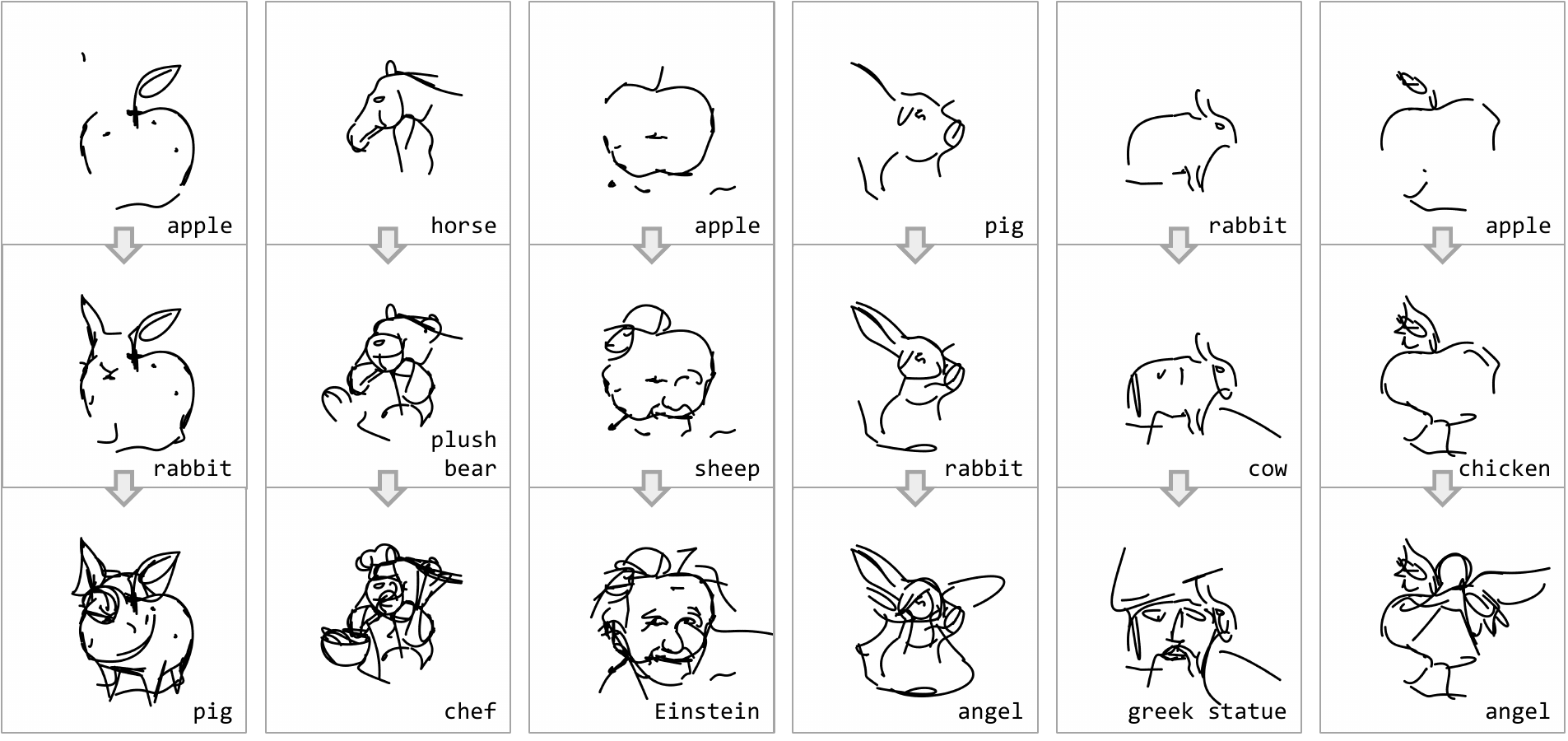}
  \caption{\textbf{Additional 3-phase progressive illusion results produced by our method.}
  }
  \label{fig:appendix_3_phase}
\end{figure*}

\begin{figure*}[t]
  \centering
  \includegraphics[width=0.89\textwidth]{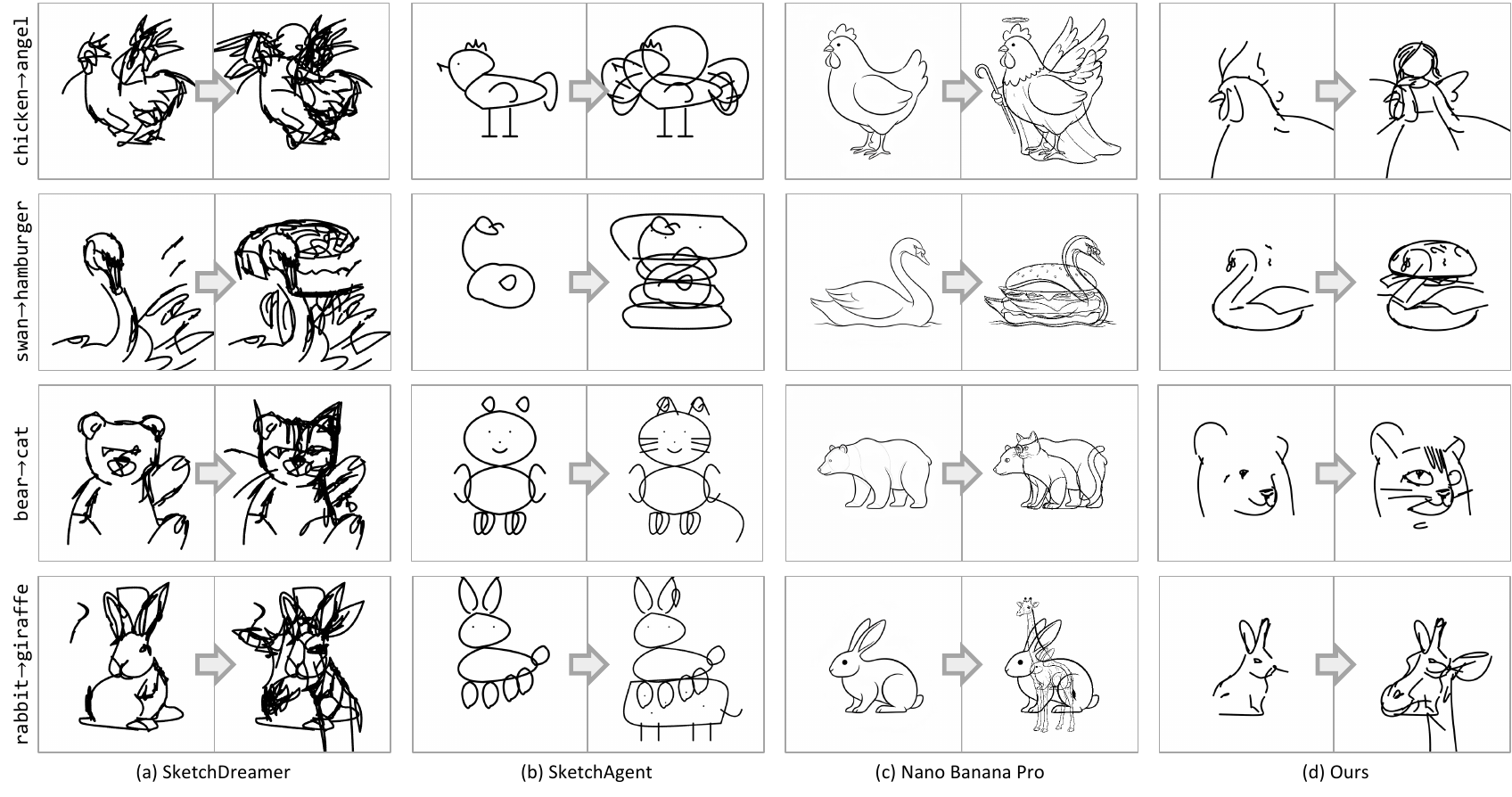}
  \caption{\textbf{Additional qualitative comparisons.}
  }
  \label{fig:appendix_qualitative_compare}
\end{figure*}

\begin{figure*}[t]
\begin{minipage}{.44\textwidth}
\centering
  \includegraphics[width=\textwidth]{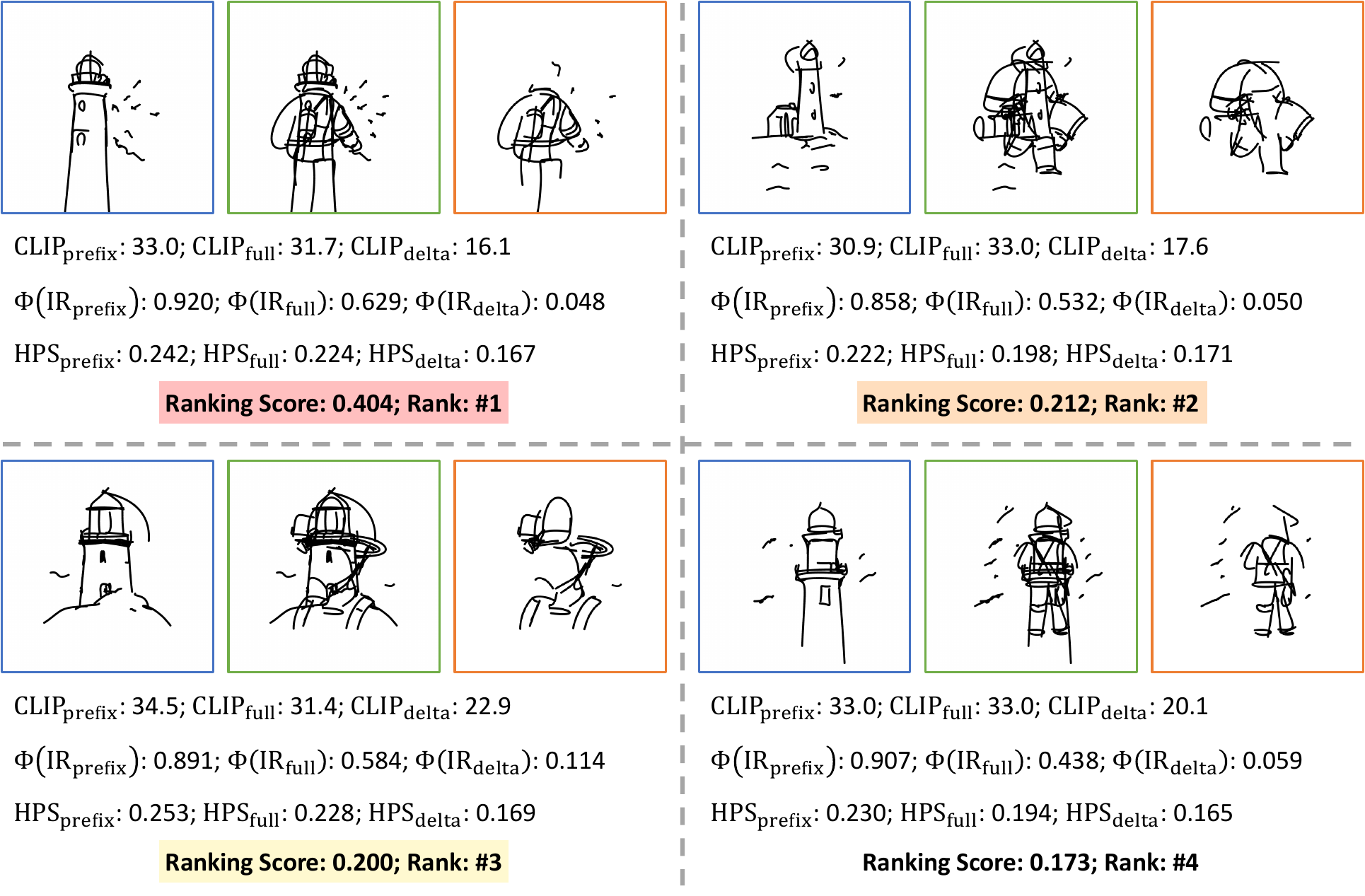}
  \caption{\textbf{Metric-based ranking.}
  }
  \label{fig:ranking_metric}
\end{minipage}
\hspace{1mm}
\begin{minipage}{.44\textwidth}
\centering
  \includegraphics[width=\textwidth]{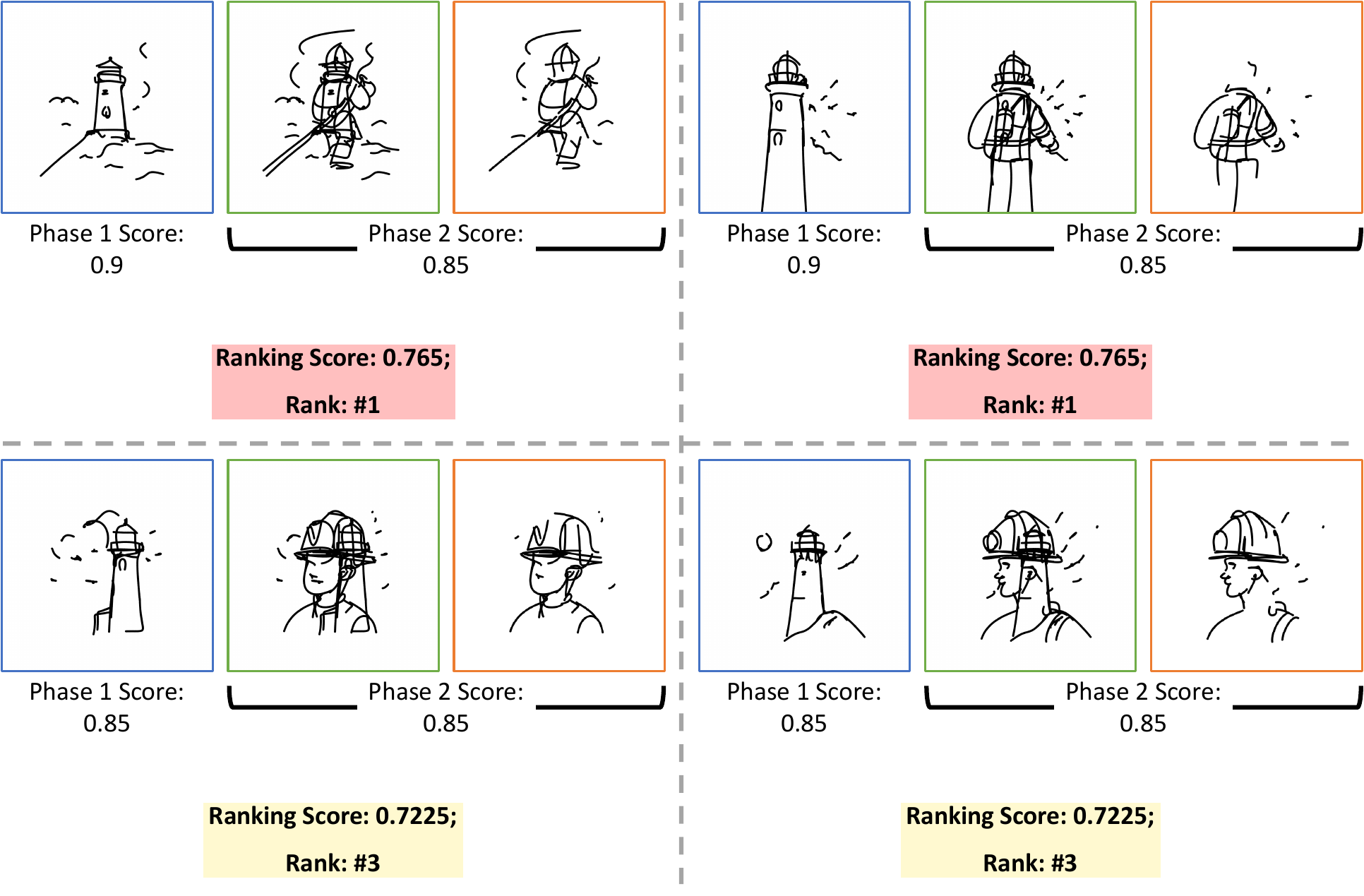}
  \caption{\textbf{GPT-based ranking.}
  }
  \label{fig:ranking_gpt}
\end{minipage}
\end{figure*}

% \begin{figure*}[t]
% \begin{minipage}{.65\textwidth}
% \centering
%   \includegraphics[width=\textwidth]{figures/appendix_bspline.pdf}
%   \caption{\textbf{Extensions on variable-width B-spline strokes.}
%   % ~\yulunliu{Change these cases so that the prompts are not duplicated.}
%   }
%   \label{fig:appendix_bspline}
% \end{minipage}
% \hspace{1mm}
% \begin{minipage}{.32\textwidth}
% \centering
%   \includegraphics[width=\textwidth]{figures/appendix_color.pdf}
%   \caption{\textbf{Extension on colored strokes.}
%   }
%   \label{fig:appendix_color}
% \end{minipage}
% \end{figure*}

\begin{figure*}[t]
\begin{minipage}{.29\textwidth}
\centering
  \includegraphics[width=\textwidth]{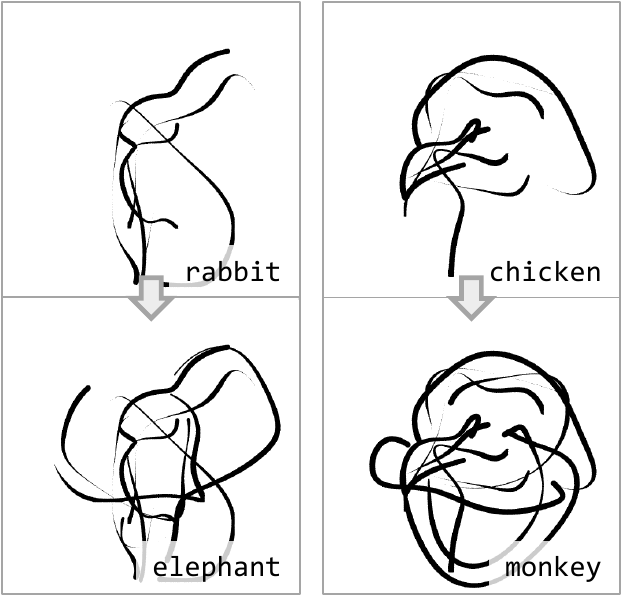}
  \caption{\textbf{Variable-width B-spline.}}
  \label{fig:appendix_bspline}
\end{minipage}
\hspace{1mm}
\begin{minipage}{.29\textwidth}
\centering
  \includegraphics[width=\textwidth]{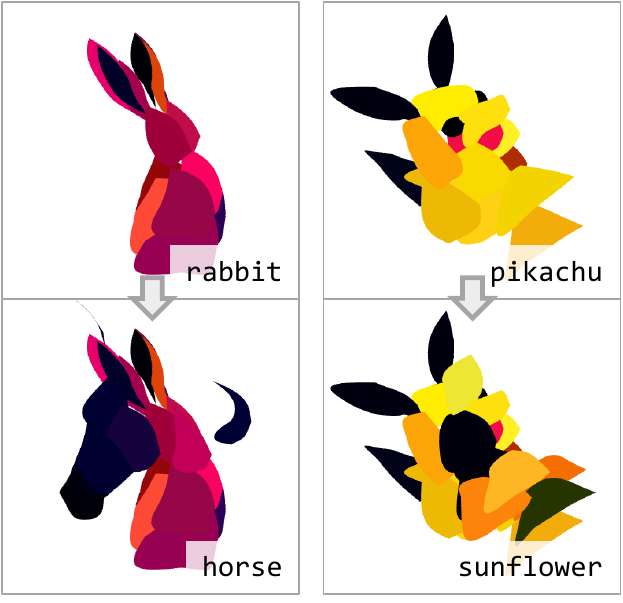}
  \caption{\textbf{Extension on vector graph.}}
  \label{fig:appendix_vector_graph}
\end{minipage}
\hspace{1mm}
\begin{minipage}{.29\textwidth}
\centering
  \includegraphics[width=\textwidth]{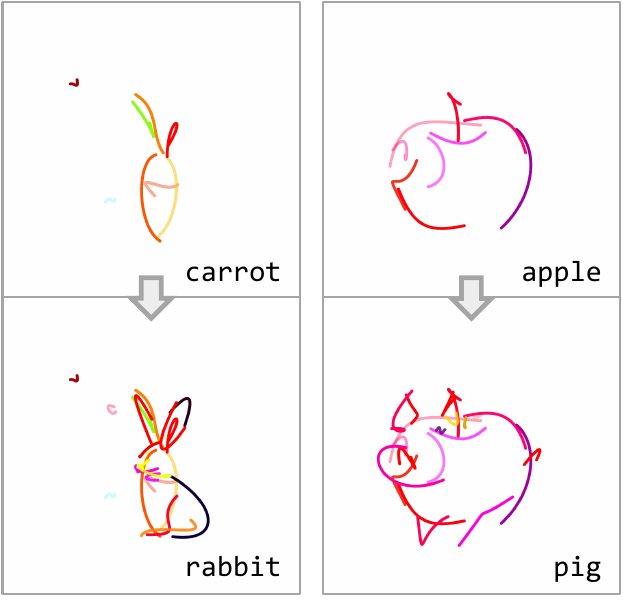}
  \caption{\textbf{Extension on colored strokes.}}
  \label{fig:appendix_color}
\end{minipage}
\end{figure*}

% \newpage
% \clearpage
% \appendix

\section{VLM Prompt Templates}
\label{sec:vlm_prompts}
\raggedbottom

This section provides the complete GPT-4o prompt templates used in our
VLM-based evaluation pipeline (Sec.~3.4 of the main paper),
corresponding to the two scoring phases illustrated in Fig.~5.
All prompts request JSON-formatted output and are queried at
temperature~$= 0$ to ensure deterministic, reproducible scoring.

% ----------------------------------------------------------------
\subsection{Phase 1 Prompt (Prefix Sketch Scoring)}
% ----------------------------------------------------------------

The Phase~1 prompt evaluates the \textbf{prefix-only sketch}
($\mathcal{S}_{\text{prefix}}$), assessing how clearly it depicts the
first target concept. Scoring priorities are, in order:
(1)~likeness to the target, (2)~recognizability as a coherent object,
and (3)~single-object integrity.

\medskip
\noindent\textbf{System Prompt:}

\begin{tcolorbox}[promptbox]
\begin{Verbatim}[fontsize=\footnotesize]
You are evaluating a single-object sketch image for a pairwise
illusion task.
\end{Verbatim}
\end{tcolorbox}

\medskip
\noindent\textbf{User Prompt}
(\texttt{\{PHASE\_LABEL\}} is replaced at runtime with the target
concept string, e.g.\ \texttt{rabbit}):

\begin{tcolorbox}[promptbox]
\begin{Verbatim}[fontsize=\footnotesize]
You are evaluating a single-object sketch image for a pairwise 
illusion task.

Target concept for this image: {PHASE_LABEL}

You must output ONE final score from 0 to 10 (can be decimals) 
that reflects overall quality for this phase.

Scoring priorities (most important first):
1) Likeness: How clearly the sketch depicts {PHASE_LABEL}.
2) Recognizability: Whether it looks like a coherent, 
   recognizable object rather than random scribbles.
Secondary consideration:
3) Single-object integrity: The image should depict one main 
   object only (not multiple separate objects).

Penalties (apply only as needed):
- Not recognizable as any coherent object: final score must be 
  <= 2.
- Multiple distinct objects or clearly separate parts: cap score 
  at <= 4.
- Depicts a different object more strongly than {PHASE_LABEL}: 
  cap score at <= 3.

Output format (STRICT JSON, no extra text):
{"final_score": <number>,
 "short_reason": "<max 18 words>"}
\end{Verbatim}
\end{tcolorbox}

% ----------------------------------------------------------------
\subsection{Phase 2 Prompt (Full Sketch vs.\ Delta Stroke Scoring)}
% ----------------------------------------------------------------

The Phase~2 prompt receives \textbf{two images simultaneously}:
the full sketch ($\mathcal{S}_{\text{full}}$, Image~1) and the delta
strokes alone ($\mathcal{S}_{\text{delta}}$, Image~2). Beyond assessing
how clearly the full sketch depicts the second target concept, the prompt
enforces an \emph{integration check} (anti-overlay criterion): the full
sketch must be meaningfully more complete than the delta strokes alone,
confirming that prefix strokes provide genuine structural scaffolding
rather than being overwritten.

\medskip
\noindent\textbf{System Prompt:}

\begin{tcolorbox}[promptbox]
\begin{Verbatim}[fontsize=\footnotesize]
You are evaluating a pairwise-illusion sketch with two images.
\end{Verbatim}
\end{tcolorbox}

\medskip
\noindent\textbf{User Prompt}
(\texttt{\{PHASE\_LABEL\}} is replaced with the second target concept,
e.g.\ \texttt{horse}; Image~1~$= \mathcal{S}_{\text{full}}$,
Image~2~$= \mathcal{S}_{\text{delta}}$):

\begin{tcolorbox}[promptbox]
\begin{Verbatim}[fontsize=\footnotesize]
You are evaluating a pairwise-illusion sketch
with two images:

Image 1: FULL sketch (phase_full) -- all strokes combined.
Image 2: DELTA sketch (phase_delta) -- only strokes added in this 
         phase.

Target concept for the final result: {PHASE_LABEL}

Your job: produce ONE final score from 0 to 10 (can be decimals)
that reflects the overall quality of the FULL sketch as the final 
phase.

Primary scoring priorities (most important):
1) Likeness on phase_full: How clearly phase_full depicts 
   {PHASE_LABEL}.
2) Recognizability on phase_full: Whether phase_full depicts a 
   coherent object. Dense or complex strokes are acceptable if 
   they form a clear structure.
3) Integration check (anti-overlay): phase_full should be 
   meaningfully better/more complete as {PHASE_LABEL} than 
   phase_delta.
   - If phase_delta alone already looks as complete as phase_full, 
   or phase_full is not clearly better, apply a penalty.

Secondary considerations (supporting, not dominant):
4) Single-object integrity: should depict one main object only
   (not two separate objects).
5) Cleanliness: avoid excessive messy strokes. If phase_full is 
   clearly recognizable as {PHASE_LABEL}, cleanliness is not a 
   major concern.

Hard constraints / caps:
- phase_full not recognizable: final score <= 5.
- phase_full wrong class: final score <= 3.
- phase_full contains two distinct objects: <= 4.
- phase_full not clearly better than phase_delta: <= 3.

Output format (STRICT JSON, no extra text):
{"final_score": <number>,
 "short_reason": "<max 20 words>",
 "integration_note": "<'full>>delta' / 'full>delta'
                       / 'similar' / 'delta>full'>"}
\end{Verbatim}
\end{tcolorbox}

% ----------------------------------------------------------------
% \subsection{Score Aggregation}
% ----------------------------------------------------------------

% Phase~1 scores directly quantify prefix sketch quality.
% Phase~2 scores capture full-sketch recognizability penalized by the
% integration check. The GPT-ranking final score is:
% \[
%   \mathcal{R}_{\text{GPT}}
%   = \text{Score}_{\text{Phase\,1}} \times \text{Score}_{\text{Phase\,2}},
% \]
% ensuring both phases must score highly for the overall ranking to be
% high, preventing a dominant phase from masking failure in the other.
% The \texttt{integration\_note} field (Phase~2 only) provides qualitative
% evidence to support manual review of each candidate's structural
% contribution.
\section{User Study Details}
\label{sec:user_study}

We conducted two user studies with 143 participants using Google Forms. Participants were aged approximately 20--50 years, and each participant completed both studies sequentially.

\subsection{Study 1: Method Comparison}
Participants were shown 10 questions, each presenting a different prompt pair. For each question, four illusion sketches labeled (A)--(D) were displayed side by side, generated by SketchDreamer~(A)\cite{qu2023sketchdreamer}, our method~(B), SketchAgent~(C)\cite{vinker2025sketchagent}, and Nano Banana Pro~(D), respectively.
Each sketch was displayed as a triplet: Phase~1 (black), Phase~2 (black), and an animated GIF toggling between Phase~1 and Phase~2 (blue), allowing participants to directly perceive the structural transition. Participants were asked:
\textit{``Which sketch best represents a good illusion sketch?''} and judged based on three criteria: (1) clear semantics at each phase, (2) smooth structural transition from Phase~1 to Phase~2, and (3) a perceptual reversal effect rather than mere stroke accumulation. Participants selected one of (A)--(D), or ``Other'' if none were satisfactory. A representative question is shown in Fig. ~\ref{fig:userstudy}~(top).

\subsection{Study 2: Ranking Pipeline Validation}
Participants were shown 4 questions, each presenting our top-4 ranked outputs for the same prompt pair. Participants were asked:
\textit{``Which of the following sketches do you consider successful illusion sketches?''} Multiple selections were allowed, including selecting none. The same three criteria were provided as guidance. This study measures whether our ranking pipeline reliably surfaces high-quality results. A representative question is shown in Fig. ~\ref{fig:userstudy}~(bottom).

\begin{figure}[t]
  \centering
  \includegraphics[width=\linewidth]{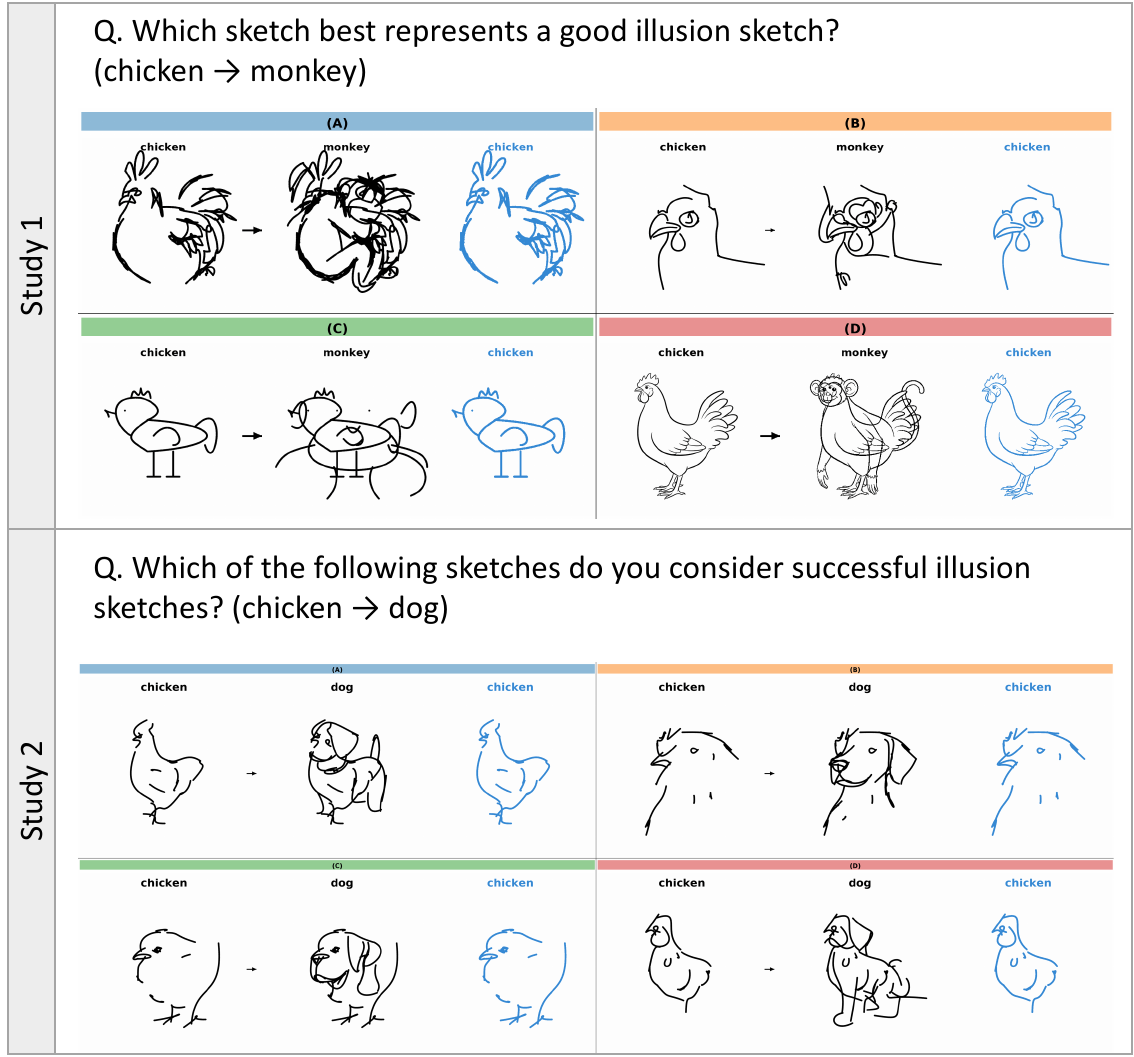}
  \caption{\textbf{Representative survey questions from Study~1 (top) and Study~2 (bottom).} In Study~1, participants selected the best illusion sketch among four methods: SketchDreamer~(A), Ours~(B), SketchAgent~(C), and Nano Banana Pro~(D). In Study~2, participants selected all sketches they considered successful from four candidates generated by our method.}
  \label{fig:userstudy}
\end{figure}
\section{Initialization and Implementation Details}
\label{sec:impl_details}

\paragraph{Stroke representation.}
Each stroke is a single-segment cubic Bézier curve with 4 control points, yielding a learnable parameter tensor of shape $N \times 4 \times 2$. Optimization is performed at a resolution of $512 \times 512$ with stroke width $2.5$\,px. The final SVG is exported at $1024 \times 1024$ with stroke width $5$\,px, preserving the stroke-to-canvas ratio.

\paragraph{Stroke initialization.}
Strokes are initialized near the canvas center (gathered strategy; see ablation Fig.~11 of the main paper). An anchor $p_0$ is sampled from $\mathcal{U}([0.3,\,0.7]^2)$ in normalized coordinates, and each subsequent control point is perturbed by $\delta \sim \mathcal{U}([-0.025,\,0.025]^2)$, giving a displacement radius of $0.05$. Points are then scaled to pixel coordinates. The random seed is fixed to $0$; minor variations may arise from CUDA nondeterminism.

\paragraph{Optimization.}
We use Adam (lr $= 0.8$, $2{,}000$ iterations). Both SDS branches share a classifier-free guidance scale of $100$ over a fully frozen Stable Diffusion v1.5 backbone. The overlay loss weight is $\lambda_{\text{overlay}} = 0.1$. Default stroke counts are $k = 16$ (prefix) and $N = 32$ (total), giving $16$ delta strokes.

\paragraph{Overlay loss.}
The spatial buffer in Eq.~(3) of the main paper is computed by applying Gaussian blur ($\sigma = 2.0$, kernel $15 \times 15$) to the separately rasterized prefix and delta maps before the normalized inner product.

\paragraph{Runtime.}
All experiments run on a single NVIDIA RTX 4090, requiring ${\sim}13$ minutes for two-phase and ${\sim}15$ minutes for three-phase illusions.
\section{Reproducibility and Variance Analysis}
\label{sec:reproducibility}

\begin{figure}[t]
  \centering
  \includegraphics[width=\linewidth]{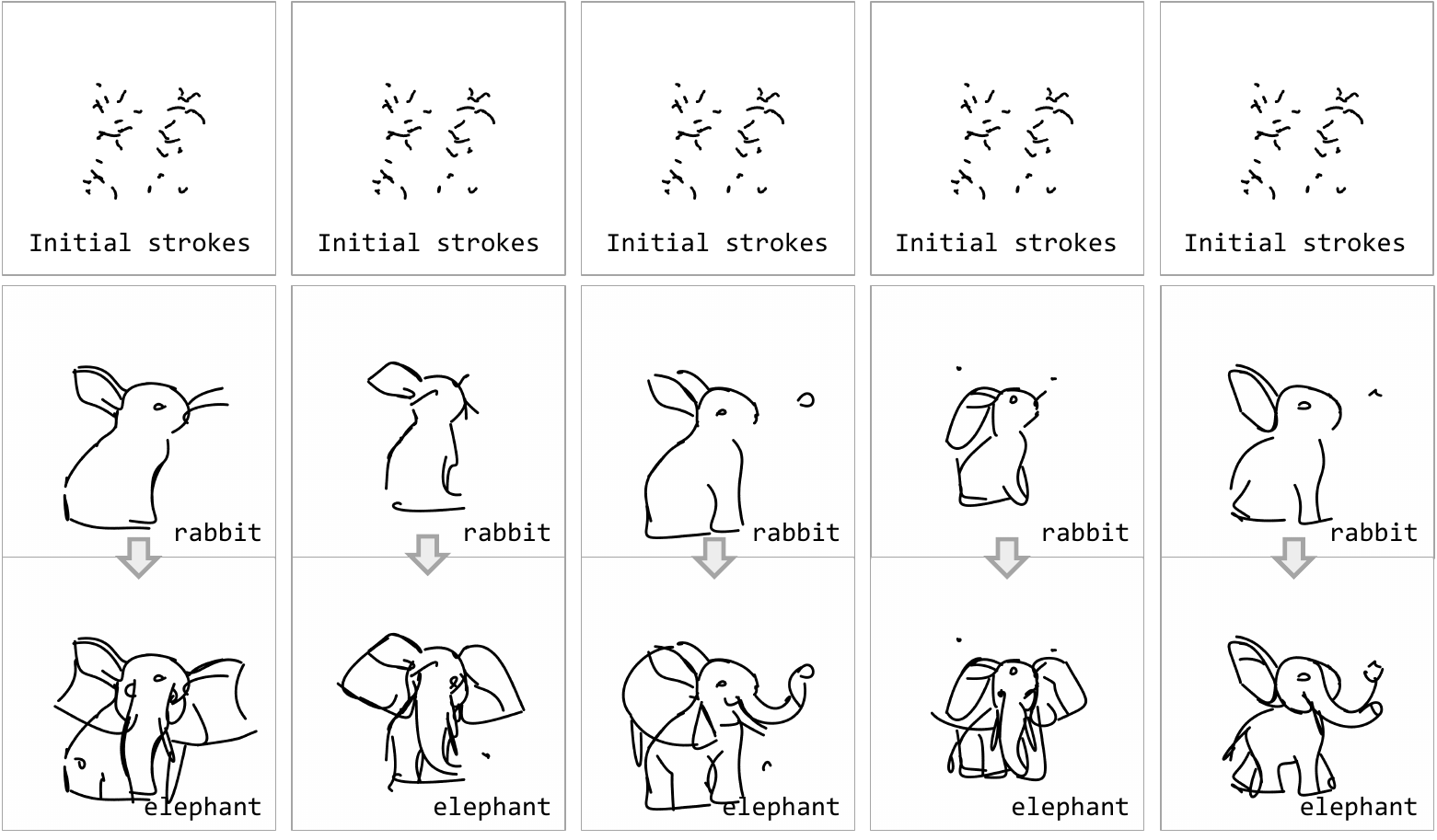}
  \caption{\textbf{Five independent runs on the same prompt pair (rabbit $\rightarrow$ elephant) with a fixed random seed.} Minor variations in stroke geometry arise from CUDA nondeterminism, but all runs yield recognizable and structurally coherent illusions at both phases.}
  \label{fig:cuda}
\end{figure}

To assess the robustness of our method under CUDA nondeterminism, we ran the same prompt pair (rabbit $\rightarrow$ elephant) five times with a fixed random seed, varying only the CUDA execution order. Figure~\ref{fig:cuda} shows all five results.
Despite minor geometric variations across runs, all five outputs are recognizable at both phases and exhibit a clear structural transition, confirming that our method produces consistently high-quality illusions under fixed initialization.

\section{Quantitative Ablation Studies}
\label{sec:supp-ablation}

We provide quantitative results for the three ablation studies discussed in Sec.~4.3 of the main paper, evaluating stroke initialization, optimization strategy, and overlay loss. Each setting is run on 5 prompt pairs with 30 illusions per setting, using identical prompts, stroke counts, and stroke widths. Results are summarized in Tab.~\ref{tab:ablation}.

\begin{table}[t]
\centering
\small
\caption{
\textbf{Ablation study.} 
We evaluate three design choices: stroke initialization 
(scattered vs.\ gathered), optimization strategy 
(sequential vs.\ joint), and overlay loss. Abl.~1 vs.~Ours 
confirms gathered initialization is critical for 
convergence; Abl.~2 vs.~3 shows joint optimization clearly 
beats sequential; Abl.~3 vs.~Ours establishes overlay loss 
as the key enabler of structural concealment. Best in 
\textbf{bold}, second \underline{underlined}.
}
\resizebox{\columnwidth}{!}{%
\begin{tabular}{l|cc|c|c|ccc|c}
\toprule
& Init & Optim. & Overlay & CLIP $\uparrow$ & \multicolumn{3}{c|}{Concealment (structural)} & $C_{\text{semantic}}$ \\
\cmidrule(lr){6-8}
& (g/s) & (j/seq) & Loss & Avg min & CLIP $\uparrow$ & IR $\uparrow$ & HPS $\uparrow$ & CLIP $\uparrow$ \\
\midrule
Abl.\ 1 & scattered & joint & \checkmark        & 28.121          & 3.759          & 1.019          & 0.033          & 0.921 \\
Abl.\ 2 & gathered  & seq   & $\times$          & 27.792          & 1.520          & 0.365          & 0.015          & \textbf{1.000} \\
Abl.\ 3 & gathered  & joint & $\times$          & \textbf{30.690} & 2.421          & 0.765          & 0.027          & \textbf{1.000} \\
\midrule
\textbf{Ours} & gathered & joint & \checkmark   & \underline{30.494} & \textbf{5.723} & \textbf{1.259} & \textbf{0.036} & \textbf{1.000} \\
\bottomrule
\end{tabular}
}
\label{tab:ablation}
\end{table}

\paragraph{Optimization Strategy.}
As shown in Tab.~\ref{tab:ablation} (Abl.~2 vs.~3), sequential optimization scores only 1.520 on concealment CLIP compared to 2.421 for joint optimization, with consistent trends observed across IR and HPS metrics. This confirms that freezing prefix strokes commits them to a rigid local minimum for Concept A, leaving delta strokes to build on an incompatible foundation. Joint optimization navigates two competing gradient fields simultaneously, discovering a common structural subspace where features serve dual roles, and is therefore essential for high-quality progressive illusion sketches.

\paragraph{Stroke Initialization.}
Tab.~\ref{tab:ablation} (Abl.~1 vs.~Ours) shows that scattered initialization scores only 3.759 on concealment CLIP compared to 5.723 for centered gathered initialization, with IR and HPS further corroborating this gap. Since our objective function is highly non-convex, spatial concentration is critical for convergence; scattered strokes fail to form the coherent spatial structure required to simultaneously serve two semantic interpretations. We therefore adopt centered gathered initialization to balance density with spatial coverage, avoiding potential boundary clipping.

\paragraph{Overlay Loss.}
Removing $\mathcal{L}_{\text{overlay}}$ drops concealment CLIP from 5.723 to 2.421 (Tab.~\ref{tab:ablation}, Abl.~3 vs.~Ours), the most significant drop among all ablated components, with IR and HPS showing similarly pronounced degradation. Without this constraint, semantic guidance alone fails to prevent spatial redundancy, causing delta strokes to clutter the prefix rather than structurally integrating with it. $\mathcal{L}_{\text{overlay}}$ enforces spatial complementarity by penalizing overlap between prefix and delta strokes, ensuring that prefix strokes serve as essential structural scaffolding for the final concept rather than being obscured. This confirms that geometric constraints are indispensable for generating clean progressive illusions.
\section{Applications: Additive and Subtractive Modes.}
\begin{figure}[t]
    \centering
    \includegraphics[width=\linewidth]{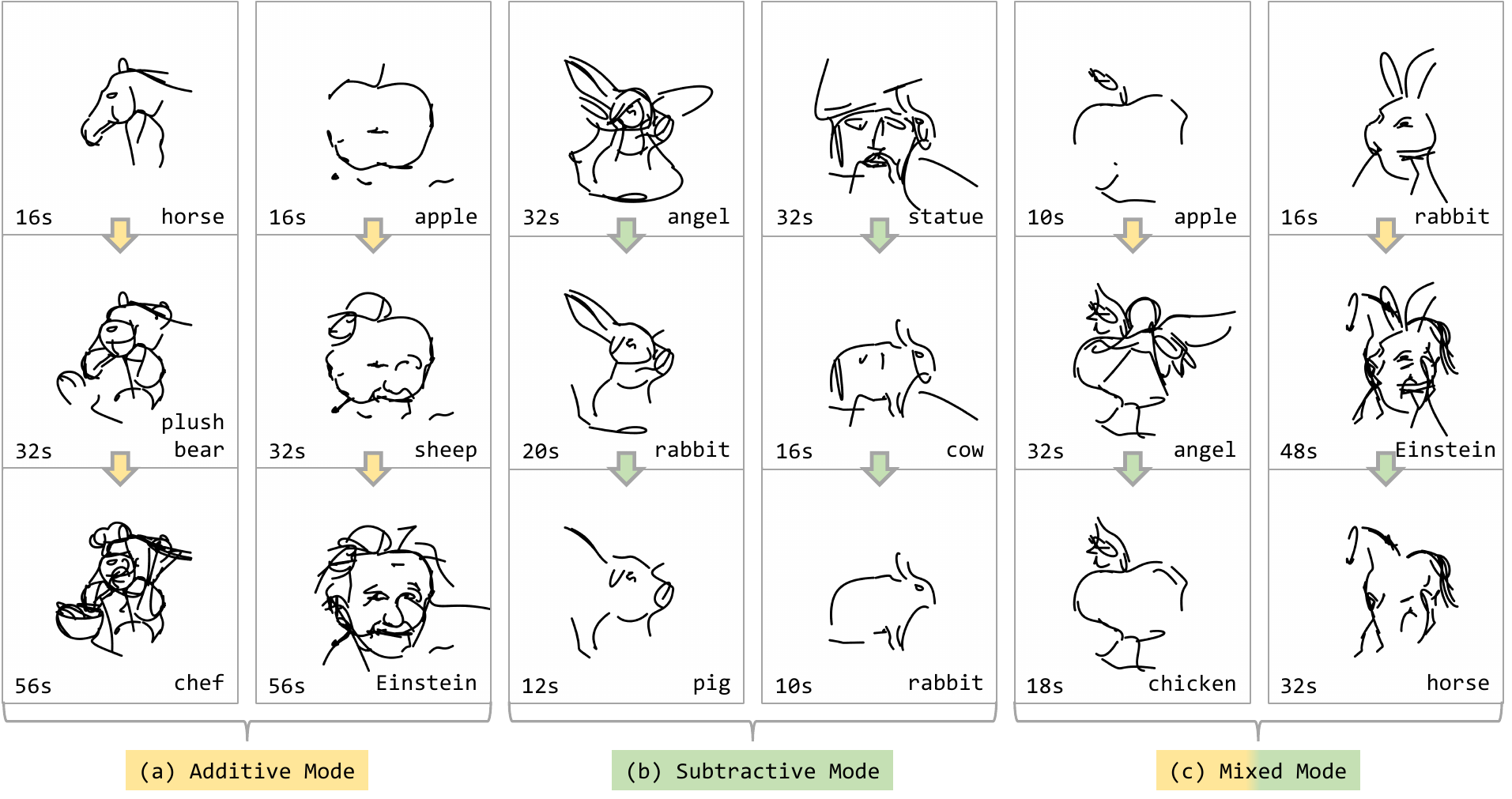}
    \caption{
        \textbf{Three interaction paradigms enabled by our framework.}
        Yellow arrows denote stroke addition; green arrows denote stroke removal.
        \textbf{(a) Additive mode:} concepts emerge through sequential stroke accumulation.
        \textbf{(b) Subtractive mode:} the full sketch is presented first; delta strokes are removed in reverse order to reveal earlier concepts, with no modification to the underlying optimization.
        \textbf{(c) Mixed mode:} addition and subtraction are interleaved within a single sequence, where the perceived concept at each step is determined by which stroke subset is rendered.
    }
    \label{fig:suppl_application}
\end{figure}

Our framework supports three interaction paradigms beyond standard 
generation (Fig.~\ref{fig:suppl_application}).
\textbf{(a) Additive mode} progressively accumulates strokes across phases.
\textbf{(b) Subtractive mode} begins from the full sketch and progressively 
removes delta strokes ($S_{\delta_2}$, then $S_{\delta_1}$) to transition 
concepts, requiring no re-optimization---only a reversal of rendering order.
\textbf{(c) Mixed mode} interleaves both directions within one sequence. For \textit{apple}$\to$\textit{angel}$\to$\textit{chicken}, adding $S_{\delta_1} \cup S_{\delta_2}$ to the prefix yields \textit{angel}, while subsequently removing $S_{\delta_2}$ recovers \textit{chicken}.
For \textit{rabbit}$\to$\textit{Einstein}$\to$\textit{horse}, a minor optimization adjustment ensures that adding $S_{\delta}$ to $S_{\text{prefix}}$ yields \textit{Einstein}, while subtracting $S_{\text{prefix}}$ instead reveals \textit{horse} as an emergent concept.
These variants demonstrate that semantic interpretation can be flexibly controlled by the rendered subset, order, or a minor reformulation of the optimization objective.

\end{document}